\documentclass[twoside,11pt]{article}
\usepackage{jair,rawfonts}

\usepackage{acronym} 
\usepackage{algorithm} 
\usepackage{algorithmicx} 
\usepackage[noend]{algpseudocode} 
\usepackage{amsmath} 
\usepackage{booktabs} 
\usepackage{cleveref} 
\usepackage{dsfont} 
\usepackage{enumitem} 
\usepackage[T1]{fontenc}  
\usepackage{longtable,tabularx} 
\usepackage{mathrsfs} 
\usepackage{pgfplots}
\usepackage{placeins}
\usepackage{siunitx}
\usepackage{subcaption} 
\usepackage{tikz}
\usepackage{tikzscale}
\usepackage[textwidth=0.75in]{todonotes}
\usepackage{txfonts}
\usepackage{url}

\usepackage[american]{babel}
\usepackage[backend=biber,natbib,bibstyle=apa,citestyle=authoryear-comp,maxcitenames=2,uniquelist=false,apamaxprtauth=99]{biblatex}
\DeclareLanguageMapping{american}{american-apa}
\DeclareFieldFormat{apacase}{#1}
\DeclareMathAlphabet{\mathcal}{OMS}{cmsy}{m}{n} 

\makeatletter 
\renewcommand{\todo}[2][]{\tikzexternaldisable\@todo[#1]{#2}\tikzexternalenable}
\makeatother

\usepackage{aircraftshapes}

\DeclareMathOperator*{\argmax}{arg\,max}

\algnewcommand{\LineComment}[1]{\State \(\triangleright\) #1}

\usetikzlibrary{external}
\usetikzlibrary{patterns}
\usetikzlibrary{shapes,arrows,fit}
\usetikzlibrary{positioning}
\usetikzlibrary{arrows.meta, calc, shapes}

\pgfplotsset{compat=newest}
\pgfplotsset{every axis legend/.append style={%
cells={anchor=west}}
}
\usepgfplotslibrary{polar}
\usepgfplotslibrary{groupplots}

\tikzset{%
			>={Latex[width=2mm,length=2mm]},
            base/.style = {rectangle, rounded corners, draw=black,
                           minimum width=1cm, minimum height=1cm,
                           text centered},
            block/.style = {base, minimum width=2.5cm, minimum height=1.5cm},
            envstyle/.style = {block, fill=red!35},
            sutstyle/.style = {block, fill=blue!10},
            simstyle/.style = {base, thick, fill=gray!15, minimum width=4cm},
            metasimstyle/.style = {base, line width=0.07cm, fill=orange!80!gray!40, minimum width=4cm},
            learnerstyle/.style = {block, fill=blue!20},
            rewardstyle/.style = {block, fill=green!70!gray!20},
            componentstyle/.style = {block, fill=blue!10},
            aircraftstyle/.style = {block, draw=none, fill=brown!50!red!30},
            casstyle/.style = {block, fill=orange!50},
            termstyle/.style = {base, thick, fill=green!20},
            nontermstyle/.style = {base, fill=blue!10},
            explainstyle/.style = {text=blue, align=left},
}
\pgfdeclarelayer{metasimlayer}
\pgfdeclarelayer{simlayer}
\pgfdeclarelayer{aircraftlayer}
\pgfsetlayers{metasimlayer,simlayer,aircraftlayer,main}

\setlist{noitemsep,topsep=0.5ex}

\newacro{acasx}[ACAS X]{Airborne Collision Avoidance System}
\newacro{apl}[JHUAPL]{Johns Hopkins University Applied Physics Laboratory}  
\newacro{ast}[AST]{adaptive stress testing}  
\newacro{atp}[ATP]{automated theorem proving}  
\newacro{dast}[DAST]{differential adaptive stress testing}
\newacro{faa}[FAA]{Federal Aviation Administration}
\newacro{fteg}[FTEG]{Fast-Time Encounter Generator}
\newacro{ga}[GA]{genetic algorithm}
\newacro{hstp}[HSTP]{hybrid systems theorem proving}
\newacro{icao}[ICAO]{International Civil Aviation Organization}
\newacro{llcem}[LLCEM]{Lincoln Laboratory Correlated Aircraft Encounter Model}
\newacro{lladm}[LLADM]{Lincoln Laboratory Aircraft Dynamics Model}
\newacro{ltl}[LTL]{linear temporal logic}
\newacro{mcts}[MCTS]{Monte Carlo tree search}
\newacro{mctspw}[MCTS-PW]{Monte Carlo tree search with progressive widening}
\newacro{mctssa}[MCTS-SA]{Monte Carlo tree search for seed-action simulators}
\newacro{mitll}[MIT-LL]{MIT Lincoln Laboratory}
\newacro{mdp}[MDP]{Markov decision process}
\newacro{nasa}[NASA]{National Aeronautics and Space Administration}
\newacro{nmac}[NMAC]{near mid-air collision}
\newacro{pmc}[PMC]{probabilistic model checking}
\newacro{pomdp}[POMDP]{partially observable Markov decision process}
\newacro{ra}[RA]{resolution advisory}
\newacro{rrt}[RRT]{rapidly-exploring random tree}
\newacro{ta}[TA]{traffic alert}
\newacro{tcas}[TCAS]{Traffic Alert and Collision Avoidance System}
\newacro{uct}[UCT]{upper confidence tree}

\acrodefplural{mdp}[MDPs]{Markov decision processes}
\acrodefplural{pomdp}[POMDPs]{partially observable Markov decision processes}
\acrodefplural{ra}[RAs]{resolution advisories}
\acrodefindefinite{mdp}{an}{a}
\acrodefindefinite{nmac}{an}{a}
\acrodefindefinite{ra}{an}{a}

\begin{document}

\title{Adaptive Stress Testing: Finding Likely\\Failure Events with Reinforcement Learning }

\author{\name Ritchie Lee \email ritchie.lee@nasa.gov\\
       \addr NASA Ames Research Center, Moffett Field, CA 94035 
       \AND
       \name Ole J. Mengshoel \email ole.j.mengshoel@ntnu.no\\
       \addr Norwegian University of Science and Technology \\
       NO-7491, Trondheim, Norway 
       \AND
       \name Anshu Saksena \email anshu.saksena@jhuapl.edu \\
       \name Ryan W. Gardner \email ryan.gardner@jhuapl.edu \\
       \name Daniel Genin \email daniel.genin@jhuapl.edu \\
       \name Joshua Silbermann \email joshua.silbermann@jhuapl.edu \\
       \addr Johns Hopkins University Applied Physics Laboratory \\ 
       11100 Johns Hopkins Rd., Baltimore, MD 20723
       \AND
       \name Michael Owen \email michael.owen@ll.mit.edu \\
       \addr MIT Lincoln Laboratory, 244 Wood St., Lexington, MA 02421
       \AND
       \name Mykel J. Kochenderfer \email mykel@stanford.edu\\
       \addr Stanford University,
       496 Lomita Mall, Stanford, CA, 94305}


\maketitle


\begin{abstract}
Finding the most likely path to a set of failure states is important to the analysis of safety-critical systems that operate over a sequence of time steps, such as aircraft collision avoidance systems and autonomous cars.  In many applications such as autonomous driving, failures cannot be completely eliminated due to the complex stochastic environment in which the system operates.  As a result, safety validation is not only concerned about whether a failure can occur, but also discovering which failures are \emph{most likely} to occur.  
This article presents \acf{ast}, a framework for finding the most likely path to a failure event in simulation.  We consider a general black box setting for partially observable and continuous-valued systems operating in an environment with stochastic disturbances.  We formulate the problem as a \acl{mdp} and use reinforcement learning to optimize it.  The approach is simulation-based and does not require internal knowledge of the system, making it suitable for black-box testing of large systems.  We present different formulations depending on whether the state is fully observable or partially observable.  In the latter case, we present a modified Monte Carlo tree search algorithm that only requires access to the pseudorandom number generator of the simulator to overcome partial observability. We also present an extension of the framework, called \acf{dast}, that can find failures that occur in one system but not in another.  This type of differential analysis is useful in applications such as regression testing, where we are concerned with finding areas of relative weakness compared to a baseline.  We demonstrate the effectiveness of the approach on an aircraft collision avoidance application, where a prototype aircraft collision avoidance system is stress tested to find the most likely scenarios of near mid-air collision.
\end{abstract}

\section{Introduction}
\label{sec:introduction}
Understanding how failures occur is important to the design, evaluation, and certification of safety-critical systems such as aircraft collision avoidance systems \parencite{Kochenderfer2012next} and autonomous cars \parencite{Bouton2018}.  The knowledge informs decisions that reduce the probability and impact of failures and prevent loss of life and property.  
We consider one of the key problems in failure analysis, which is finding the most likely sequence of transitions from an initial state to a failure state.  That is, we aim not only to find a scenario that results in a failure event, but also to maximize the probability of the scenario that causes it.  For example, one might be interested in the most likely scenario where an autonomous vehicle collides with another vehicle given probabilistic models of sensor noise and other vehicles.  Maximizing probability is important because in many domains failures can almost always be reached.  For example, another car approaching us head-on can swerve into our lane at the last moment, leading to a collision we could not avoid.  However, not all failures are equally likely to occur. Incorporating a probabilistic model can significantly improve the relevance of the failure examples uncovered.

The problem is challenging in many ways.  Many failure events of interest, such as an autonomous car colliding with a pedestrian, cannot be analyzed by considering the system alone. Because failures occur as a result of sequential interactions between the system and its environment, failure analysis must be performed over the combined system.  Systems that operate in large, continuous, and stochastic environments thus present modeling and scalability challenges to analysis.   The problem is exacerbated because search occurs over a sequence of time steps, which results in an exponential number of possible futures.  Exhaustive consideration of all possible paths is generally intractable.  In addition to a large search space, failure states can also be extremely rare and difficult to reach, which is generally the case for mature safety-critical systems.

\subsection{Related Work} 
Existing methods for finding failure events can be broadly separated into two categories.  Formal verification constructs a mathematical model of the system and rigorously proves or exhaustively checks whether a safety property holds \parencite{DSilva2008,Kern1999}.  The properties are expressed using a formal logic, such as \ac{ltl} \parencite{Pnueli1977}.
Probabilistic model checking (PMC)\acused{pmc} is a formal verification method that verifies properties over stochastic models with discrete states, such as Markov chains and probabilistic timed automata \parencite{Katoen2016,Essen2016,Gardner2016}.  
\ac{pmc} exhaustively evaluates properties over all states and paths subject to probabilistic constraints.  Algorithms can prune infeasible paths to reduce the search space \parencite{Katoen2016}.
Automated theorem proving (ATP)\acused{atp} uses computer algorithms to automatically generate mathematical proofs \parencite{Gallier2015}.  Systems and assumptions are modeled in formal logic, and then \ac{atp} is applied to prove whether a property holds \parencite{Kouskoulas2017}.  If a proof is generated successfully, then that property holds over the entire model.  However, if the algorithm fails to generate a proof, then it is uncertain whether the property holds.  Hybrid systems theorem proving (HSTP)\acused{hstp} is a variant of \ac{atp} based on differential dynamic logic, which is a real-valued, first-order dynamic logic for hybrid systems \parencite{Jeannin2015}.  A hybrid system model can capture both continuous and discrete dynamic behavior.  The continuous behavior is described by a differential equation and the discrete behavior is described by a state machine or automaton.
Formal verification methods can provide a counterexample when a property does not hold.  More importantly, these methods provide completeness guarantees over the entire model, i.e., they can prove the absence of violations.  The major challenge is scalability.  Due to their exhaustive nature, they have difficulty scaling to systems with large and complex state spaces.  

The second category of methods relies on sampling, which trades completeness in favor of scalability.  These methods rely on the availability of a simulator.  Simulation models have very few requirements.  They only require the ability to draw samples of the next state.  As a result, they can contain large sophisticated models and directly embed software systems.  Scenarios can be manually crafted by a domain expert or they can sweep over a low-dimensional parametric model \parencite{Chludzinski2009}.  An alternative approach is to run simulations using a stochastic model of the system's operating environment \parencite{Kochenderfer2010,Holland2013}.  Sequences of states are sampled from the simulator and then the sequences are checked for failures.  Because sampling does not optimize for failures, it can take a very large number of simulations to encounter the correct sequence and combination of stochastic values to encounter a failure state.  
Importance sampling and the cross-entropy method have been used to accelerate the discovery of rare events \parencite{Kim2016,OKelly2018,Zhao2016}.  However, while this approach improves upon direct Monte Carlo sampling, it does not leverage the sequential structure of the problem.  

The problem of finding failure examples directly, known as \emph{falsification}, has been considered in the literature.  Falsification does not exhaustively cover the space as in verification, but instead uses search and optimization techniques to find failures.  As a result, falsification methods can scale to much larger systems, but generally cannot prove the absence of failures.  One approach formulates the problem as a minimization of a robustness measure \parencite{stlrobustness}.
Global optimization algorithms, such as simulated annealing and Nelder-Mead, have been applied, for example in the tool S-TaLiRo \parencite{Annapureddy2011}.  Global optimization methods are not aware of the temporal relationship between optimization variables and thus these methods do not leverage the sequential structure of the problem during optimization.
An alternative approach formulates the problem as a trajectory planning problem and optimizes using variants of \acp{rrt} \parencite{Dreossi2015}.  The \ac{rrt} approach involves growing a tree from an initial state to a failure state by sampling a random point in the state space and growing the tree towards that target point.  In the limit of infinite samples, the tree can reach any point in the reachable search space.  However, the procedure requires evaluating the closeness of two states to determine which node in the tree to expand.  In its original context, \ac{rrt} was applied to trajectory planning in physical spaces, which are Euclidean.  When the dimensionality of the state space is large and contains variables with mixed types and scales, it is unclear what distance metric to use.  The search also requires directly operating on the state, which cannot be applied to simulators with hidden state.   
Recent work has also proposed using reinforcement learning \parencite{Akazaki2018} and tree search methods \parencite{falstar}, which are similar to the approach in this article.  The key difference between existing falsification work and the current work is that we solve a slightly different problem.  Existing falsification algorithms aim to find the failure example with the lowest robustness.  This article aims to find the most likely failure example given a probabilistic disturbance model.  
A second difference is that we introduce an abstraction and learning algorithm in this article that can be a applied to (non-Markovian) systems with hidden states and partially unknown disturbance distributions.

\subsection{Our Approach} 
This article presents \acf{ast}, a method for finding the most likely path to a failure event.  We consider a simulator that is a Markov process with discrete time and continuous state.  \ac{ast} adaptively guides the sampling of paths to optimize finding failure events and maximizing the path likelihood.  It also leverages the sequential structure of the problem for optimization.  As a result, it can scale to much larger problems and efficiently search for the most likely failure paths.  Our \ac{ast} method formulates the search problem as a sequential decision process and then applies reinforcement learning algorithms to optimize it.  We present formulations for both fully observable systems, where the algorithm has full access to the simulator state, and partially observable systems, where some or all of the simulator states are hidden.  In the latter case, we present a modified \acf{mcts} algorithm, called \acf{mctssa} that only requires access to the pseudorandom number generator of the simulator to overcome partial observability. \ac{ast} treats the simulator as a black box, where the system transition behavior is not known.  As a result, the approach can be applied to a broad range of systems.
We base our algorithm on \ac{mcts} \parencite{Kocsis2006} because of its ability to scale to large problems and because it can be easily modified to handle non-Markovian systems.  Other algorithms can be used as well.  For example, deep reinforcement learning has been used for \ac{ast} to analyze the safety of autonomous cars \parencite{Koren2018}.  Failure scenarios found by \ac{ast} can then be further analyzed to extract common patterns, e.g., by using an automated categorization algorithm \parencite{Lee2018b}.

In some applications, it may also be valuable to evaluate failure paths not in absolute terms, but in relation to a baseline system.  That is, we are not interested in cases where both systems fail, but rather cases where the test system fails but the baseline system does not.   
We call this type of analysis \emph{differential stress testing}.  Such situations arise, for example, during regression testing where a new version of a system is compared to a previous one to identify areas of comparative weakness. 
One way to compare the relative behavior of two systems is to evaluate them against a common set of scenarios.  For example, scenarios can be generated by running Monte Carlo simulations using the test system, and then replaying them on the baseline system \parencite{Holland2013}.  However, this approach suffers from the same inefficiency issues as before.  The size and complexity of the state space, the rarity of failures, and the exponential explosion of searching over sequences make encountering failure events extremely unlikely.  In fact, the issue is even more pronounced in differential stress testing because the failure event needs to occur in the system under test but not the baseline. 
We present an extension of \ac{ast} to the differential setting called \acf{dast}.  The approach finds the most likely path to a failure event that occurs in the system under test, but not in the baseline system.  \ac{dast} follows the same general formulation as \ac{ast}.  However, in the differential setting, we search two simulators in parallel and maximize the difference in the outcomes of the simulators.  

\subsection{Case Study: ACAS X} 
We demonstrate the effectiveness of \ac{ast} and \ac{dast} for stress testing the next-generation \ac{acasx} \parencite{Kochenderfer2012next}.  \ac{acasx} has been recently accepted as the next international standard for aircraft collision avoidance. The system replaces the previous system, called \ac{tcas}, which has performed very well in the past, but is not optimized for the next-generation airspace \parencite{Kuchar2007}. For example, the number of nuisance alerts is expected to dramatically increase with the rising density of air traffic.  This article describes work performed while \ac{acasx} was under development by the \ac{faa}.  As part of the \ac{acasx} validation team, we obtained various prototypes of \ac{acasx} from the \ac{faa} and stress tested them in simulated aircraft encounters to find the most likely scenarios of \ac{nmac}.  Our experiments include single-threat (two-aircraft) encounters, multi-threat (three-aircraft) encounters, and differential stress testing against \ac{tcas}.  Our results were reported to the \ac{acasx} development team to inform development and assess risk.  We highlight the main findings from these reports, along with the general methods we introduced.


The main contributions of this article are summarized as follows:
\begin{itemize}
    \item We present \acf{ast}, a novel framework that formulates finding the most likely failure scenario as a sequential decision-making problem.  The formulation enables reinforcement learning solvers to be applied. 
    \item We present an abstraction for \ac{ast} that uses pseudorandom seeds to overcome partial observability in the testing simulator.
    \item We present \acf{dast}, an extension of \ac{ast} for finding the most likely scenarios where a failure occurs in the system under test, but not in a baseline system. 
    \item We present a case study of the \ac{acasx} aircraft collision avoidance system, where we characterize its most likely scenarios of near mid-air collisions (NMACs) both in absolute terms and relative to the existing \ac{tcas}. 
\end{itemize}

The remaining sections are organized as follows. \Cref{sec:background} reviews sequential decision processes and \ac{mcts}.  \Cref{sec:ast} presents an overview of the \ac{ast} framework, followed by formulations of \ac{ast} for fully observable and partially observable systems.  The section also presents the \ac{mctssa} algorithm for optimizing partially observable systems.  \Cref{sec:dast} presents \ac{dast}, an extension of \ac{ast} to differential stress testing.  Finally, \Cref{sec:application} presents the results of analyzing near mid-air collisions in an aircraft collision avoidance system.



\section{Background}
\label{sec:background}

In this section, we first present a brief overview of sequential decision process, which is the mathematical framework underlying \ac{ast}. We then present an algorithm for solving a sequential decision process, called \acl{mcts}.

\subsection{Sequential Decision Process}
A sequential decision process models a situation where an agent makes a sequence of decisions in an environment to maximize a reward function \parencite{dmubook}.  If the environment is known and its state is fully observable, then the problem can be formulated as a Markov decision process (MDP)\acused{mdp}. 
\Iac{mdp} is a 5-tuple $\langle S,A,P,R,\gamma \rangle$, where $S$ is a set of states; $A$ is a set of actions; $P$ is the transition probability function, where $P(s' \mid s,a)$ is the probability of choosing action $a \in A$ in state $s \in S$ and transitioning to next state $s' \in S$; and $R$ is the reward function, where $R(s,a)$ is the reward for taking $a$ in $s$.  We define the transition function $T$ for convenience, where $T(s,a)$ samples the next state $s'$ from the distribution $P(s' \mid s,a)$.  The parameter $\gamma \in [0,1]$ is the discount factor that governs how much to discount the value of future rewards.

In \iac{mdp}, the agent chooses an action $a = \pi(s)$ according to its policy $\pi$.  The system evolves probabilistically to the next state $s' \sim T(s,a)$. The agent then receives reward $r = R(s,a)$ for the transition.  The assumption that the transition function depends only on the current state and action is known as the \emph{Markov property}.
In cases where the underlying process is Markovian, but the agent only observes part of the state, the problem is \iac{pomdp} \parencite{dmubook}.  At each time step, the agent observes an observation $o \in O$, which depends probabilistically on state $s$ and action $a$.  The actions in \iac{pomdp} at time $t$ can only be based on the history of observations up to time $t$.  

A \emph{simulation} iteratively samples the sequential decision process to produce a \emph{path}, which is a sequence of states and actions (and observations in the case of \iac{pomdp}).
In this article, we assume that the model is \emph{episodic} in that it terminates in a finite number of steps.  The \emph{terminal time} $t_{\text{\text{end}}}$ is the first time the state enters a terminal state or reaches a maximum number of time steps $t_{\text{max}}$.  We assume that once the simulation terminates, the agent receives the terminal reward and does not collect any additional rewards thereafter.  Because we consider a finite number of steps, we set $\gamma=1$.  Moreover, we define the \emph{return} $G$ to be the sum of rewards collected over a path, i.e., $G=\sum_{t=0}^{t_{\text{end}}}{r_t}$.  

Reinforcement learning algorithms can be used to optimize sequential decision problems through sampling of the transition function $T$ \parencite{Sutton1998,Wiering2012}.  The learners adapt their sampling strategy during the search, enabling them to efficiently search large and complex state spaces.  
Model-free value-based reinforcement learning algorithms are a class of learning algorithms that aim to estimate the \emph{state-action value function} $Q(s,a)$, which is the expected sum of rewards resulting from choosing action $a$ in state $s$ and following an optimal policy $\pi^*(s)$ thereafter.  An \emph{optimal action} $a^*$ is an action that maximizes the state-action value function at state $s$, i.e., $a^*=\argmax_{a}{Q(s,a)}$. An \emph{optimal policy} $\pi^*(s)$ is a function that gives an optimal action $a^*$ for each state $s$.  The objective of reinforcement learning algorithms is to find an optimal policy $\pi^*(s)$.

\subsection{Monte Carlo Tree Search}
\acf{mcts} is a state-of-the-art heuristic search algorithm for optimizing sequential decision processes \parencite{Kocsis2006,Browne2012}.  \ac{mcts} incrementally builds a search tree using a combination of directed sampling based on estimates of the state-action value function and undirected sampling based on a fixed distribution, called \emph{rollouts}.  
During the search, sampled paths from the simulator are used to incrementally estimate $Q(s,a)$ and the optimal policy at nodes in the tree.  To account for the uncertainty in the estimates, which may lead to premature convergence, \ac{mcts} encourages exploration by adding an exploration term to the state-action value function to encourage choosing paths that have not been explored as often.  The exploration term optimally balances the selection of the best action estimated so far with the need for exploration to improve the quality of current value estimates.  By doing so, \ac{mcts} adaptively focuses the search towards more promising areas of the search space.  The effect of the exploration term diminishes as the number of times a state is visited increases.

This article uses a variant of \ac{mcts} called \acf{mctspw}, which extends \ac{mcts} to large or continuous action spaces \parencite{Coulom2007,Chaslot2008a}.  When the action spaces are large (or infinite), visited actions are not revisited sufficiently through sampling alone, which hinders the quality of value estimates.  The benefit of the progressive widening in \ac{mctspw} is that it forces revisits to existing nodes and slowly allows new nodes to be added as the total number of visits increases.  Progressive widening, sometimes also called progressive unpruning, stabilizes value estimates and prevents explosion of the branching factor of the tree.  \ac{mctspw} converges asymptotically to the optimal solution as the number of iterations increases \parencite{Couetoux2011}.

\section{Adaptive Stress Testing}
\label{sec:ast}

Adaptive stress testing (AST)\acused{ast} aims to find the most likely path from a start state to a failure state in a discrete-time simulator \parencite{Lee2015}.  The overall approach is to formulate the search as a sequential decision-making problem and then use reinforcement learning to optimize it.

We consider a \emph{simulator} $\mathcal{S}$ that contains a system under test (or simply \emph{system}) $\mathcal{M}$ interacting with an \emph{environment} $\mathcal{E}$.  The system with state $\mu \in M$ takes action $a \in A$ based on observation $o \in O$ of the environment state $z \in Z$.  The system interacts with the environment over discrete time $t \in [0,\dots,t_{\text{end}}]$.  The simulator state $s \in S$ is the stacked system and environment states $[\mu,z]$.  We use subscript to denote the variable at time $t$ and subscript colon range to denote the sequence of a variable over a range of time steps.  For example, the simulator state path up to time $t$ is $s_{0:t}=[s_0,\dots,s_t]$.  The state and action of the system depend on the environment observations and are modeled by 

\begin{equation}
    \mu_{t+1},a_t = \mathcal{M}({o}_{0:t})
    \label{eq:sut_po}
\end{equation}
The environment state and observation evolve over time depending on the actions of the system and disturbances $x \in X \subseteq \mathbb{R}^n$, where $n$ is the dimensionality of the disturbances.  The term disturbances is very broad and encompasses any stochastic variables that can influence the environment.  For example, disturbances can control the magnitude and direction of the wind, or they can control other actors in the environment such as pedestrians and other vehicles.

\begin{equation}
   z_{t+1},o_{t+1} = \mathcal{E}({a}_{0:t},{x}_{0:t}) 
    \label{eq:env_po}
\end{equation}
We assume the disturbances are independent across time and distributed with probability density $p(x \mid s)$.  
The disturbance model can be constructed through expert knowledge or learned from data.
We define an \emph{event space} $E \subset S$ where the event of interest occurs.  While this article focuses on failure events, an event can be arbitrarily defined.  We use the notation $s \in E$ to indicate that an event has occurred.  Alternatively, we may use the Boolean variable $e$ to indicate whether an event has occurred.  

The goal of \ac{ast} is to find the \emph{most likely failure path}, which is the path with the highest likelihood subject to the constraint that the final state is an event.

\begin{equation}
    \begin{aligned}
    & \underset{x_0,\dots,x_{t_{\text{end}}}}{\max}
    & & {\prod_{t=0}^{t_{\text{end}}-1} p(x_{t} \mid s_t)}\\
    & \text{subject to}
    & & s_{t_{\text{end}}} \in E
    \end{aligned}
\label{eq:ast_obj}
\end{equation}

\begin{figure}[!htbp]
\centering
\resizebox{0.6\columnwidth}{!}
{
\begin{tikzpicture}[node distance=1.5cm, every node/.style={font=\large}, align=center]
	\node (env) [envstyle, xshift=-2cm, yshift=1cm] {Environment $\mathcal{E}$};
    \node (sut) [sutstyle, right of=env, xshift=1.75cm, yshift=0cm] {System\\Under Test $\mathcal{M}$};
    
    \begin{pgfonlayer}{simlayer}
        \node (sim) [fit={($(env.north west)+(-2mm,7mm)$) ($(sut.south east)+(2mm,-2mm)$)},simstyle]{};
    	\node at ($(sim.north west)+(1.4cm,-4mm)$) [font={\large}] {Simulator $\mathcal{S}$};
    \end{pgfonlayer}
    \node (learner) [learnerstyle, below of=sim, yshift=-2cm] {Reinforcement\\Learner $\mathcal{A}$};
    
    \draw[->,transform canvas={xshift=-0.5cm}] (learner) -- (sim) node[midway,left]{disturbance $x$};
    \draw[->,transform canvas={xshift=0.5cm}] (sim) -- (learner) node[midway,right,align=left]{sim state $s$,\\reward $r$};
	\node (arrow) [single arrow, single arrow head extend=0.18cm, draw=black, fill=gray, minimum width=1cm, minimum height=1.4cm,right=0.4cm of sim] {};
    \node [right=0.3cm of arrow,align=left] {\textbf{Most Likely}\\\textbf{Failure Path}};
\end{tikzpicture}
}
\caption{Adaptive stress testing.  A reinforcement learning agent chooses disturbances in the environment to be most adversarial to the system under test.  The reward function is crafted to search for failure events and maximize transition likelihoods.}
\label{fig:stress_testing}
\end{figure}
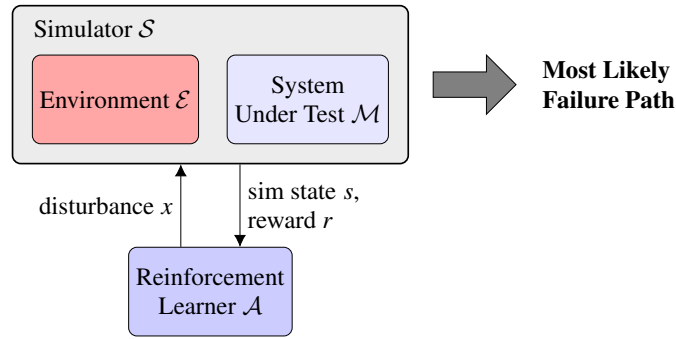

\ac{ast} formulates the search as a reinforcement learning problem by considering a reinforcement learning agent $\mathcal{A}$ that acts as an adversary to the system under test.  We let the agent choose disturbances so that the environment is as challenging to the system under test as possible. 
\Cref{fig:stress_testing} illustrates the general \ac{ast} concept.  The simulator models the behavior of the system under test and the environment.  The simulator is treated as a black box by the agent.  At each time step, the agent observes the simulator state, chooses a disturbance, and receives a reward.  Through repeated interactions with the simulator, the agent learns to choose disturbances that maximize the reward it receives.  By choosing a reward function that rewards failure events and higher likelihood transitions, the agent learns to optimize for the most likely failure path.


\subsection{Full Observability}
\label{sec:ast-fo}
If the simulator state is fully observable and the behavior of $\mathcal{M}$ and $\mathcal{E}$ depend only on the input at the current time, then the transition equations become:

\begin{eqnarray}
    \mu_{t+1}, a_t &=& \mathcal{M}(\mu_t,o_{t})\label{eq:sut_fo}\\
   z_{t+1},o_{t+1} &=& \mathcal{E}(z_{t},a_{t},x_{t}) \label{eq:env_fo}
\end{eqnarray}

\begin{figure}[!htbp]
\centering
\resizebox{0.9\columnwidth}{!}
{
\begin{tikzpicture}[node distance=1.5cm,
    every node/.style={font=\large},
    align=center]
	\node (env) [envstyle, xshift=-4.25cm, yshift=1cm] {Environment $\mathcal{E}$};
    \node (sut) [sutstyle, right of=env, xshift=4.25cm, yshift=0cm] {System\\Under Test $\mathcal{M}$};
    
    \begin{pgfonlayer}{simlayer}
        \node (sim) [fit={($(env.north west)+(-4mm,9mm)$) ($(sut.south east)+(4mm,-4mm)$)},simstyle]{};
    	\node at ($(sim.north west)+(1.4cm,-3.2mm)$) [font={\large}] {Simulator $\mathcal{S}$};
    \end{pgfonlayer}
    
    \draw[->,transform canvas={yshift=0.2cm}] (sut.west) -- (env.east) node[midway,above]{action $a$} ;
    \draw[->,transform canvas={yshift=-0.2cm}] (env.east) -- (sut.west) node[midway,below]{observation $o$};
    
    \node (learner) [learnerstyle, right=3.25cm of sim] {Reinforcement\\Learner $\mathcal{A}$};
    \node (reward) [rewardstyle, below=of learner, xshift=-2.5cm] {Reward\\Function $R$};
    
    \draw[->] (learner) -- (sim) node[midway,above]{state $s$,\\disturbance $x$};
    \draw[<-] (learner.south) |- (reward.east) node[pos=0.35,right]{reward\\$r$} ;
    \draw[<-] (reward.north) |- (learner);
    \draw[->] (sim.west) -- ++(-0.75cm,0) -- ++(0,-4.25cm)node[xshift=1.5cm,yshift=0.5cm]{next state $s'$} -- ++(+17cm,0) |- (learner);
\end{tikzpicture}
}
\caption{Adaptive stress testing of a fully observable system.  The problem is modeled as \iac{mdp} where the agent observes the simulator state and chooses a disturbance at each time step.  Traditional reinforcement learning algorithms can be applied to solve the \ac{mdp}.}
\label{fig:astblock}
\end{figure}
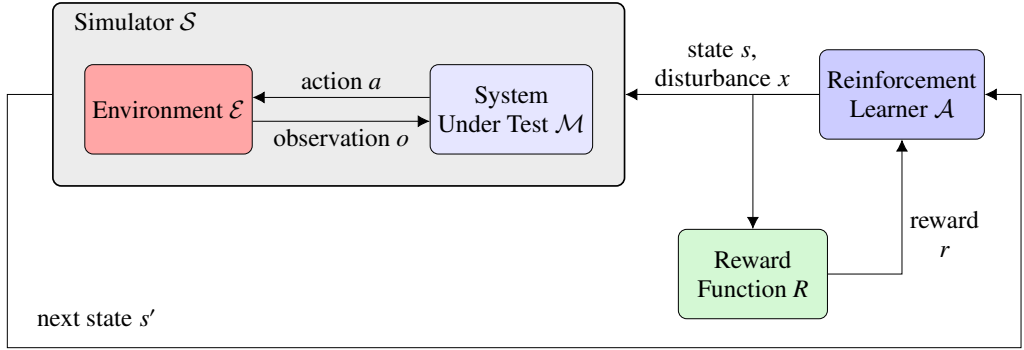

We formulate the \ac{ast} problem as \iac{mdp} as follows.  The state of the \ac{mdp} is the state $s$ of the simulator.  The agent observes the state $s$ and chooses disturbance $x$.  The transition to the next state is given by the transition behavior of the simulator $\mathcal{S}$, which consists of the combined behavior of $\mathcal{M}$ and $\mathcal{E}$ according to Equations~\ref{eq:sut_fo} and \ref{eq:env_fo}.  The reward function $R$ is crafted to optimize Equation~\ref{eq:ast_obj}, which searches for the most likely failure path.  We describe the reward function in the following subsection.
We set $\gamma=1$ to properly account for the transition likelihood in the reward function.  
\Cref{fig:astblock} illustrates the \ac{ast} framework for the fully observable case.

\paragraph{\textbf{Reward Function.}}  The reward function is designed to find failure events as the primary objective and maximize the path likelihood as a secondary objective.  Let $R_E \in \mathbb{R}_{\geq 0}$ be the event reward, $d \in \mathbb{R}_{\geq 0}$ be the miss distance, and $E \subset S$ be the event states.  Then the reward function is given by:

\begin{equation}
	\label{eq:astmdp_reward}
	R(s,x)=
    \begin{cases}
		R_E & \text{if $s$ is terminal and $s \in E$}\\
		-d & \text{if $s$ is terminal and $s \notin E$}\\
        \log (p(x \mid s)) & \text{otherwise}
    \end{cases}
\end{equation}

The first term of Equation~\ref{eq:astmdp_reward} gives a non-negative constant reward $R_E$ if the path terminates and a failure event occurs.  If the path terminates and an event does not occur, the second term of Equation~\ref{eq:astmdp_reward} penalizes the learner by assigning the negative of the miss distance to the learner.  The miss distance $d$ is some measure defined by the user that depends on $s$ and indicates how close the simulation came to a failure.  If such a measure is not available, then $-d$ can be set to a large negative constant.  However, providing an appropriate miss distance can greatly accelerate the search by giving the learner the ability to distinguish the desirability of two paths that do not contain failure events.  
The third term of Equation~\ref{eq:astmdp_reward} maximizes the overall path likelihood by awarding the log likelihood of each transition.  Recall that reinforcement learning maximizes the expected sum of rewards.  By choosing a reward of the log likelihood at each step, the reinforcement learning algorithm then maximizes the sum of the log likelihoods, which is equivalent to maximizing the product of the likelihoods.  

Maximizing the reward function in Equation~\ref{eq:astmdp_reward} also maximizes the \ac{ast} objective in Equation~\ref{eq:ast_obj}.  The first two terms in the reward function apply at the terminal time $t_{\text{end}}$ and incentivize the agent to satisfy the constraint $s_{t_{\text{end}}} \in E$ in Equation~\ref{eq:ast_obj}.  The third term of the reward function maximizes the path likelihood.  We show the relation as follows:
\begin{equation*}
    \begin{aligned}
    \underset{x_0,\dots,x_{t_{\text{end}}}}{\max}
    & G\\
    = \underset{x_0,\dots,x_{t_{\text{end}}}}{\max}
    & \sum_{t=0}^{t_{\text{end}}}{R(s_{t},x_t)}\\
    = \underset{x_0,\dots,x_{t_{\text{end}}}}{\max}
    & \left[\sum_{t=0}^{t_{\text{end}}-1}{\log(p(x_t \mid s_t))} + R_E \cdot \mathds{1}{\{s_{t_{\text{end}}} \in E}\} - d \cdot \mathds{1}{\{s_{t_{\text{end}}} \notin E\}}\right]\\
    = \underset{x_0,\dots,x_{t_{\text{end}}}}{\max}
    & \left[\prod_{t=0}^{t_{\text{end}}-1}{p(x_t \mid s_t)} + R_E \cdot \mathds{1}{\{s_{t_{\text{end}}} \in E}\} - d \cdot \mathds{1}{\{s_{t_{\text{end}}} \notin E\}}\right]\\
    \end{aligned}
\label{eq:ast_obj_equiv}
\end{equation*}
where we have used the convexity of the logarithm function to replace the sum with a product in the last line.
The indicator function $\mathds{1}{\{b\}}$ returns $1$ if $b$ is true and $0$ otherwise.
If the difference between $R_E$ and $-d$ is sufficiently large, the learner will be incentivized to first satisfy $s_{t_{\text{end}}} \in E$ to replace the miss distance penalty with the event reward, and then maximize the path likelihood.  We do not distinguish between varying degrees of a failure.  Once we have found a failure event, all optimization effort is spent towards maximizing the path likelihood.   

The \ac{mdp} can be optimized using standard reinforcement learning algorithms \parencite{Sutton1998,Wiering2012}, which only require sampling of the transitions.  Existing reinforcement learning algorithms such as \ac{mcts} \parencite{Kocsis2006,Chaslot2008a} and Q-Learning \parencite{Watkins1992} can be applied to optimize the decision process and find the optimal path. 

\subsection{Partial Observability}
\label{sec:ast-po}

Many simulators simply do not allow access to all or any state information. For example, the collision avoidance system we analyze in \Cref{sec:application} was provided as a software binary and maintained hidden state over function calls.
We introduce an abstraction that relaxes the need for the simulator to expose its underlying state and disturbance.  First, instead of explicitly representing and passing the state into and out of the simulator as in the fully observable case, we now assume that the simulator maintains state internally and the state is updated in-place.  In other words, we have previously assumed that the simulator is stateless, but now we assume that the simulator is stateful and that the state is hidden from the learner.  Second, rather than passing disturbance values $x$ as input, we pass a pseudorandom seed $\bar{x}$ as a proxy.
A \emph{pseudorandom seed}, or just \emph{seed}, is a vector of integers used to initialize a pseudorandom number generator.  We assume that all random processes in the simulator are derived from the pseudorandom number generator and seed, so that the result of sampling from these processes is deterministic given the seed.
The simulator uses $\bar{x} \sim \mathbb{U}_{\text{seed}}$ to seed an internal random process that samples $x \sim p(x \mid s)$.  Setting the seed makes the sampling process deterministic and thus a particular seed $\bar{x}$ is deterministically tied to a particular sample of disturbance $x$.  

\subsubsection{Seed-Action Simulator}
A \emph{seed-action simulator} $\mathcal{\bar{S}}$ is a stateful simulator that uses a pseudorandom seed input to update its state in-place.  The state $s$ is not exposed externally, making the simulator appear non-Markovian to external processes, such as the reinforcement learner.  The simulator uses $\bar{x}$ to draw a sample of the disturbance $x$ and transition to the next state $s'$.  The next state replaces the current state in-place.  The simulator returns the transition likelihood $\rho = p(x \mid s)$; a Boolean indicating whether an event occurred $e = (s \in E)$; and the miss distance $d$.  While the state cannot be observed or set, the simulator transitions are deterministic given the pseudorandom seed $\bar{x}$.  This property allows a previously visited state to be revisited by replaying the sequence of pseudorandom seeds $\bar{x}_{0:t-1} = [\bar{x}_{0},\dots,\bar{x}_{t-1}]$ that leads to it starting from the initial state.  
The seed-action simulator $\mathcal{\bar{S}}$ exposes the following simulation control functions: 

\begin{itemize}
\item $\textsc{Initialize}(\mathcal{\bar{S}})$ resets the simulator $\mathcal{\bar{S}}$ to a deterministic initial state $s_0$.  The simulation state is modified in-place.
\item $\textsc{Step}(\mathcal{\bar{S}},\bar{x})$ advances the state of the simulator by pseudorandom sampling.  First, the pseudorandom seed $\bar{x}$ is used to set the state of the simulator's pseudorandom process.  Second, a sample $x \sim p(x \mid s)$ is drawn.  Third, the simulator evaluates the values of the transition likelihood $\rho$, whether the event occurred $e$, and the miss distance $d$ of the current state.  Then, the simulator transitions to the next state $s'$ using the simulator transition functions in Equations~\ref{eq:sut_po} and \ref{eq:env_po} replacing $s$ with $s'$. The simulator returns ($\rho$, $e$, $d$). 
\item $\textsc{IsTerminal}(\mathcal{\bar{S}})$ returns true if the current state of the simulator is terminal and false otherwise.  The simulator terminates if an event occurs, i.e., $s \in E$, or if the simulation has reached a maximum number of time steps $t_{\text{max}}$.
\end{itemize}

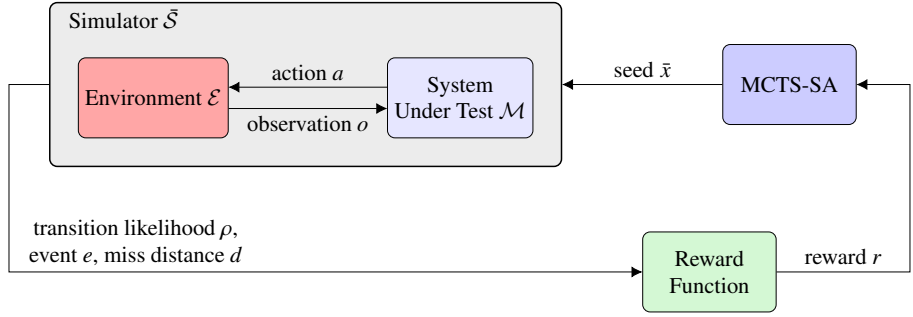
\begin{figure}[!htbp]
\centering
\resizebox{0.8\columnwidth}{!}
{
\begin{tikzpicture}[node distance=1.5cm,
    every node/.style={font=\large},
    align=center]
  
	\node (env) [envstyle, xshift=-4.25cm, yshift=1cm] {Environment $\mathcal{E}$};
    \node (sut) [sutstyle, right of=env, xshift=4.25cm, yshift=0cm] {System\\Under Test $\mathcal{M}$};
    
    \begin{pgfonlayer}{simlayer}
        \node (sim) [fit={($(env.north west)+(-4mm,9mm)$) ($(sut.south east)+(4mm,-4mm)$)},simstyle]{};
    	\node at ($(sim.north west)+(1.4cm,-3.2mm)$) [font={\large}] {Simulator $\mathcal{\bar{S}}$};
    \end{pgfonlayer}
    
    \draw[->,transform canvas={yshift=0.2cm}] (sut.west) -- (env.east) node[midway,above]{action $a$} ;
    \draw[->,transform canvas={yshift=-0.2cm}] (env.east) -- (sut.west) node[midway,below]{observation $o$};
    
    
    \node (learner) [learnerstyle, right=3cm of sim] {MCTS-SA};
    \node (reward) [rewardstyle, below=of learner, xshift=-1.5cm, yshift=-0.5cm] {Reward\\Function};
    
    \draw[->] (learner) -- (sim) node[midway,above]{seed $\bar{x}$};
    \draw[->] (reward.east) -- ++(2.5cm,0cm) node[midway,above]{reward $r$} |- (learner.east);
    \draw[->] (sim.west) -- ++(-0.75cm,0cm) |- (reward) node[pos=0.6,above]{transition likelihood $\rho$,\\event $e$, miss distance $d$};
\end{tikzpicture}
}
\caption{Adaptive stress testing of a partially observable system. We use a seed-action simulator model where the simulator retains state, but does not expose it.  The learner sets the pseudorandom seed that controls the pseudorandom number generator of the simulator making transitions deterministic.  The simulator outputs the transition likelihood of the current transition, whether an event occurred, and a miss distance.}
\label{fig:astblock2}
\end{figure}

\Cref{fig:astblock2} illustrates the \ac{ast} framework under the pseudorandom seed abstraction, where the simulator has been replaced by a seed-action simulator $\mathcal{\bar{S}}$, which maintains a hidden state.  The simulator takes a seed input $\bar{x}$ and transitions state internally.  The simulator outputs the transition likelihood $\rho$, a Boolean indicating whether an event occurred $e$, and the miss distance $d$.  The reward function translates the simulator outputs into a reward $r$.  The optimization algorithm learns from the reward to optimize over its seed inputs.  

The seed-action abstraction also provides significant practical benefits.  Large software simulators are often written in a distributed and modular fashion where each component maintains its own state.  The simulator may consist of many of these components.  Since the simulator state is the concatenation of the states of all the individual components, explicitly assembling and handling the state can break modularity and be a major implementation inconvenience.  This abstraction, which uses in-place state update and pseudorandom seeds as a proxy to the actions, alleviates the need to explicitly form the state and enables the simulator to maintain modularity of the components.  For implementation, all that is required is the ability to set the global pseudorandom seed of the simulator, which is generally easy to do in software simulators.

\subsubsection{Reward Function}
The reward function for \ac{ast} of a seed-action simulator is given in Equation~\ref{eq:poastreward}.  The reward is expressed as a function of seed-action simulator outputs and a Boolean variable $\tau$ which indicates whether the simulator has terminated, i.e., $\tau=\textsc{IsTerminal}(\mathcal{\bar{S}})$.  The three components mirror those in Equation~\ref{eq:astmdp_reward}.  

\begin{equation}
	\label{eq:poastreward}
	R(\rho,e,d,\tau)=
    \begin{cases}
		R_E & \text{if $\tau \wedge e$}\\
		-d & \text{if $\tau \wedge \neg e$}\\
        \log \rho & \text{otherwise}
    \end{cases}
\end{equation}

\subsubsection{MCTS-SA Algorithm}
We present Monte Carlo tree search for seed-action simulators (MCTS-SA)\acused{mctssa}.  We base the algorithm on the progressive widening variant of \ac{mcts} because the actions of the reinforcement learner are now pseudorandom seeds, which are vast \parencite{Coulom2007,Chaslot2008a}.
The transition behavior of the simulator is deterministic given a pseudorandom seed input.  Consequently, only a single next state is possible and there is no need to limit the number of next states.  Parameters $k$ and $\alpha$ are used for the progressive widening of actions.
The action space is the space of all pseudorandom seeds.  Since the seeds are discrete and do not have any semantic relationship, there is no need to distinguish between them.  We choose the rollout policy and the action expansion function of \ac{mcts} to uniformly sample over all seeds $\bar{x} \sim \mathbb{U}_{\text{seed}}$. Sampling seeds uniformly generates $x \sim p(x \mid s)$ in the simulator.  Because we are sampling from a continuous distribution, the samples of $x$ will be unique. 
The hidden state $s$ of the simulator is not available to the reinforcement learner. However, since the simulator is deterministic given the seed input, we can revisit a previous state by replaying the sequence of seeds that leads to it starting from the initial state.  As a result, we use the sequence of seeds $[\bar{x}_{0},\dots,\bar{x}_{t-1}]$ as the state $\bar{s}_t$ in the algorithm.  
The path with the highest path return may, and likely will, be from a path that is encountered during a rollout.  Since rollouts are not individually recorded in the tree, information about the best path can be lost.  To ensure that the algorithm returns the best path seen over the entire search, we explicitly track the highest return seen $G^*$ and the corresponding seed sequence $\bar{s}^*$.
The \ac{mctssa} algorithm is shown in \Cref{alg:ast_mcts}.  The algorithm takes as input a seed-action simulator $\mathcal{\bar{S}}$ and returns the most likely failure path represented by its seed sequence $\bar{s}^*$.  The algorithm consists of a main loop that repeatedly performs forward simulations of the system while building the search tree and updating the state-action value estimates.  The search tree $\mathcal{T}$ is initially empty.  

Each simulation runs from initial state to terminal state.   The path is determined by the sequence of pseudorandom seeds chosen by the algorithm, which falls into three stages for each simulation:

\begin{algorithm} 
\caption{\label{alg:ast_mcts}MCTS for seed-action simulators}
\begin{algorithmic}[1]
\LineComment{Inputs: Seed-action simulator $\mathcal{\bar{S}}$}
\LineComment{Returns: Seed sequence $\bar{s}^*$ that induces path with highest return $G^*$}
\Function{MCTS-SA}{$\mathcal{\bar{S}}$}
\State \textbf{global} $\bar{s}_{\text{end}} \gets \emptyset$
\State $(\bar{s}^*, G^*) \gets (\emptyset, -\infty)$
\Loop
	\State $\bar{s} \gets \emptyset$
	\State $\Call{Initialize}{\mathcal{\bar{S}}}$
	\State $G \gets \Call{Simulate}{\mathcal{\bar{S}}, \bar{s}}$
	\If{$G > G^*$}
    	\State $(\bar{s}^*, G^*) \gets (\bar{s}_{\text{end}}, G)$
    \EndIf
\EndLoop
\State \Return $\bar{s}^*$ 
\EndFunction

\Function{Simulate}{$\mathcal{\bar{S}}, \bar{s}$}
\If{$\bar{s} \not\in \mathcal{T}$} 
    \State $\mathcal{T} \gets \mathcal{T} \cup \{\bar{s}\}$  \Comment{Expansion, add new node}
	\State $(N(\bar{s}), \bar{X}(\bar{s})) \gets (0, \emptyset)$
	\State \Return $\Call{Rollout}{\mathcal{\bar{S}}, \bar{s}}$	
\EndIf
\State $N(\bar{s}) \gets N(\bar{s}) + 1$
\If{$|\bar{X}(\bar{s})| < k N(\bar{s})^\alpha$}  \Comment{Progressive widening of seeds}
    \State $\bar{x} \sim \mathbb{U}_{\text{seed}}$  \Comment{New seed}
	\State $(N(\bar{s}, \bar{x}), Q(\bar{s},\bar{x})) \gets (0,0)$
	\State $\bar{X}(\bar{s}) \gets \bar{X}(\bar{s}) \cup \{\bar{x}\}$
\EndIf
\State $\bar{x} \gets \argmax_x Q(\bar{s}, x) + c \sqrt{\frac{\log N(\bar{s})}{N(\bar{s},x)}}$  \Comment{UCT selection criterion}
\State $\tau \gets \Call{IsTerminal}{\mathcal{\bar{S}}}$
\State $(\rho, e, d) \gets \Call{Step}{\mathcal{\bar{S}}, \bar{x}}$  \Comment{Deterministic transition}
\State $r \gets \Call{Reward}{\rho, e, d, \tau}$
\If{$\tau$}
	\State $\bar{s}_{\text{end}} \gets \bar{s}$
	\State \Return $r$
\EndIf	
\State $\bar{s}' \gets [\bar{s}, \bar{x}]$
\State $q \gets r + \Call{Simulate}{\mathcal{\bar{S}}, \bar{s}'}$
\State $N(\bar{s},\bar{x}) \gets  N(\bar{s}, \bar{x}) + 1$
\State $Q(\bar{s}, \bar{x}) \gets Q(\bar{s}, \bar{x}) + \frac{q - Q(\bar{s}, \bar{x})}{N(\bar{s}, \bar{x})}$  \Comment{Update estimate of $Q$}
\State \Return $q$
\EndFunction

\Function{Rollout}{$\mathcal{\bar{S}}, \bar{s}$}
\State $\bar{x} \sim \mathbb{U}_{\text{seed}}$  \Comment{Sample seeds uniformly}
\State $\tau \gets \Call{IsTerminal}{\mathcal{\bar{S}}}$
\State $(\rho, e, d) \gets \Call{Step}{\mathcal{\bar{S}}, \bar{x}}$
\State $r \gets \Call{Reward}{\rho, e, d, \tau}$
\If{$\tau$}
	\State $\bar{s}_{\text{end}} \gets \bar{s}$
	\State \Return $r$
\EndIf
\State $\bar{s}' \gets [\bar{s}, \bar{x}]$
\State \Return $r + \Call{Rollout}{\mathcal{\bar{S}}, \bar{s}'}$
\EndFunction
\end{algorithmic}
\end{algorithm}

\begin{itemize}
\item \emph{Search.} In the search stage, which is implemented by \textsc{Simulate} (line 13), the algorithm starts at the root of the tree and recursively selects a child to follow.  At each visited state node, the progressive widening criteria (line 19) determines whether to choose amongst existing seeds or to expand the number of children by sampling a new seed.  The criterion limits the number of seeds at a state $\bar{s}$ to be no more than polynomial in the total number of visits to that state \parencite{Chaslot2008a}.  Specifically, a new seed $\bar{x}$ is sampled from a discrete uniform distribution over all seeds $\mathbb{U}_{\text{seed}}$ if $|\bar{X}(\bar{s})| < kN(\bar{s})^\alpha$, where $k$ and $\alpha$ are parameters, $\bar{X}(\bar{s})$ is the set of previously applied seeds from state $\bar{s}$, $|\bar{X}(\bar{s})|$ is the cardinality of $\bar{X}(\bar{s})$, and $N(\bar{s})$ is the total number of visits to state $\bar{s}$.  Otherwise, the existing action that maximizes 
\begin{equation}
Q(\bar{s},\bar{x}) + c \sqrt{\frac{\log N(\bar{s})}{N(\bar{s},\bar{x})}}
\label{eq:bandit}
\end{equation}
is chosen (line 23), where $c$ is a parameter that controls the amount of exploration in the search, and $N(\bar{s},\bar{x})$ is the total number of visits to seed $\bar{x}$ in state $\bar{s}$.  Equation~\ref{eq:bandit} is the \ac{uct} equation \parencite{Kocsis2006}.  The second term in the equation is an \emph{exploration bonus} that encourages selecting seeds that have not been tried as frequently.  The seed is used to advance the simulator to the next state and the reward is evaluated.  The search stage continues in this manner until the system transitions to a state that is not in the tree.

\item \emph{Expansion.} Once we have reached a state that is not in the tree $\mathcal{T}$, we create a new node for the state and add it (line 14).  The set of previously applied seeds from this state $\bar{X}(\bar{s})$ is initially empty and the number of visits to this state $N(\bar{s})$ is initialized to zero.

\item \emph{Rollout.} Starting from the state created in the expansion stage, we perform a \emph{rollout} that repeatedly samples state transitions until the desired termination is reached (line 17).  In the \textsc{Rollout} function (line 35), state transitions are drawn from the simulator with seeds chosen according to a rollout policy, which we set to sampling from $\mathbb{U}_{\text{seed}}$.
\end{itemize}

At each step in the simulation, the reward function is evaluated and the reward is used to update estimates of the state-action values $Q(\bar{s}, \bar{x})$ (line 33).  The values are used to direct the search.  At the end of each simulation, the best return $G^*$ and best path $\bar{s}^*$ are updated (lines 10--11). 
Simulations are run until the stopping criterion is met.  The criterion is a fixed number of iterations for all our experiments except for the performance study (\Cref{sec:performance}) where we used a fixed computational budget.  The algorithm returns the path with the highest return represented as a sequence of pseudorandom seeds.  The sequence of seeds can be used to replay the simulator to reproduce the failure event.

\subsubsection{Computational Complexity}
Each iteration of the \ac{mcts} main loop simulates a path from initial state to terminal state.  As a result, the number of calls to the simulator is linear in the number of loop iterations.  The computation time thus varies as $O(N_{\text{loop}} \cdot (T_{\textsc{Initialize}} + N_{\text{steps}} \cdot T_{\textsc{Step}}))$, where $N_{\text{loop}}$ is the number of loop iterations, $T_{\textsc{Initialize}}$ is the computation time of \textsc{Initialize}, $N_{\text{steps}}$ is the average number of steps in the simulation, and $T_{\textsc{Step}}$ is the computation time of the \textsc{Step} function.

\section{Differential Adaptive Stress Testing}
\label{sec:dast}
The previous section presented \ac{ast}, which can be used to find the most likely failure path in a system.  In some applications, it may also be valuable to identify failure scenarios not in absolute terms, but compared to another system---that is, areas where the system is \emph{relatively} weak compared to a baseline system.  In other words, we are not interested in the cases where both systems perform poorly, but rather where the system under test performs poorly but the baseline system performs well.  This analysis may arise, for example, when comparing two candidate solutions to determine which one may be more desirable for release.  Another use case is for regression testing where a new version of a system is compared to a previous one to see whether any new issues have been introduced.  

One way to compare the behavior of two systems is to evaluate them against a common set of testing scenarios.  For example, testing inputs can be randomly drawn using Monte Carlo from a stochastic model of the system's operating environment.  Then, the inputs can be applied to both systems and the scenarios where failure occurs in the system under test but not in the baseline system are kept. 
While this method can generate failures, the undirected nature of this approach can be very inefficient due to the size and complexity of the state space, and the rarity of failure events.  Moreover, the method does not find the most likely path to a failure, which is very valuable in the analysis of failure events.
Another, perhaps slightly better, approach is to use a stress testing method, such as \ac{ast}, to identify failure scenarios in the system under test, and then replay the scenario on the baseline system.  If the scenario does not fail on the baseline system, then accept the scenario.  This method improves upon the first method in that the failure scenarios on the system under test are being optimized.  However, the optimization process does not take into account the behavior of the baseline system.

\subsection{Approach}
We present a stress testing method, called \acf{dast}, that extends the \ac{ast} framework to the differential analysis setting while retaining its desirable properties, including scalability, efficiency, and support for black-box systems.  \ac{dast} finds the most likely path to a failure event that occurs in the system under test, but not in the baseline system \parencite{Lee2018a}.  The key idea behind \ac{dast} is to drive two simulators in parallel and maximize the difference in their outcomes.  To achieve this, we craft a new reward function that accepts the outputs of two simulators and encourages failures in one simulator but not the other.  For optimization, we regard the two parallel simulators as a larger combined simulator, then we follow the \ac{ast} approach to formulate stress testing as a sequential decision-making problem and optimize it using reinforcement learning.  One of the core advantages of the seed-action formulation introduced previously is that it uses control of the pseudorandom seed to abstract the internal state and transition behavior from the optimization procedure.  By composing two seed-action simulators into a combined simulator that is also a seed-action simulator, we can apply the \ac{mctssa} algorithm described in \Cref{alg:ast_mcts} for optimization without any modification.

\begin{figure}[!htbp]
\centering
\resizebox{0.7\columnwidth}{!}
{
\begin{tikzpicture}[node distance=1.5cm, every node/.style={font=\large}, align=center]
	\node (env1) [envstyle, xshift=-2cm, yshift=1cm] {Environment};
    \node (sut1) [sutstyle, right of=env1, xshift=1.25cm, yshift=0cm] {System\\Under Test};
    \begin{pgfonlayer}{simlayer}
        \node (sim1) [fit={($(env1.north west)+(-2mm,7mm)$) ($(sut1.south east)+(2mm,-2mm)$)},simstyle]{};
    	\node at ($(sim1.north west)+(1.4cm,-4mm)$) [font={\large}] {Simulator $\mathcal{\bar{S}}^{(1)}$};
    \end{pgfonlayer}
	\node (env2) [envstyle, below=1.5cm of env1] {Environment};
    \node (sut2) [sutstyle, right of=env2, xshift=1.25cm, yshift=0cm] {Baseline};
    \begin{pgfonlayer}{simlayer}
        \node (sim2) [fit={($(env2.north west)+(-2mm,7mm)$) ($(sut2.south east)+(2mm,-2mm)$)},simstyle]{};
    	\node at ($(sim2.north west)+(1.4cm,-4mm)$) [font={\large}] {Simulator $\mathcal{\bar{S}}^{(2)}$};
    \end{pgfonlayer}
    \begin{pgfonlayer}{metasimlayer}
        \node (sim) [fit={($(sim1.north west)+(-10mm,2mm)$) ($(sim2.south east)+(5mm,-2mm)$)},metasimstyle]{};
    	\node [font={\Large}, above=0cm of sim] {\textbf{Combined Simulator $\mathcal{\bar{S}}$}};
    \end{pgfonlayer}
    \node (reward) [rewardstyle, right=3.5cm of sim,minimum height=5cm] {Reward\\Function};
    \node (learner) [learnerstyle, above=1.75cm of reward,xshift=-4.5cm] {MCTS-SA};
    
    \draw[->] (sim1.east) -- (sim1-|reward.west) node[pos=0.6,above]{likelihood $\rho^{(1)}$,\\event $e^{(1)}$,\\miss distance $d^{(1)}$};
    \draw[->] (sim2.east) -- (sim2-|reward.west) node[pos=0.6,above]{likelihood $\rho^{(2)}$,\\event $e^{(2)}$,\\miss distance $d^{(2)}$};
    \draw[->] (reward.north) |- (learner.east) node[pos=0.7,above]{reward $r$};
    \draw[->] (learner) -- ($(learner-|sim.west)+(0.5cm,0)$) node(seedpoint)[pos=0.8,above] {seed $\bar{x}$} |- (sim1.west);
    \draw[->] (learner) -- ($(learner-|sim.west)+(0.5cm,0)$) |- (sim2.west);
\end{tikzpicture}
}
\caption{Differential adaptive stress testing framework.  Two parallel simulators, one running the system under test and one running the baseline, are searched simultaneously.  The search drives one simulator to failure while keeping the second simulator away from one.}
\label{fig:dast}
\end{figure}
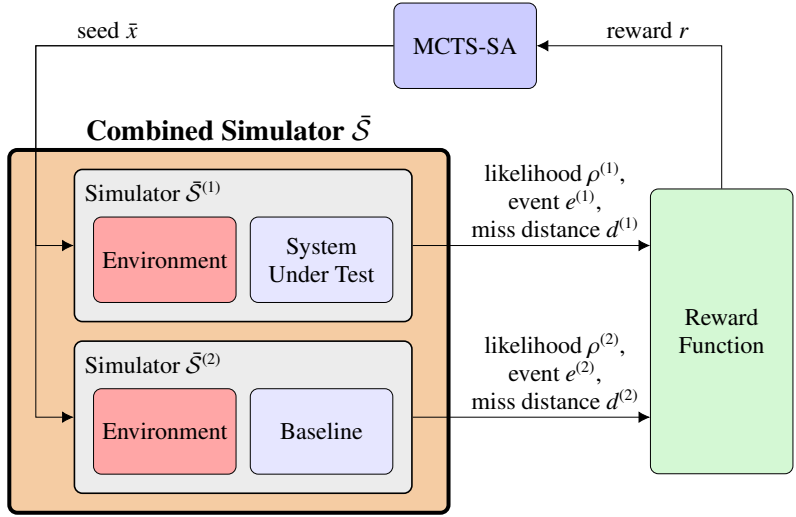

\Cref{fig:dast} illustrates the \ac{dast} framework.  We create two instances of the simulator, $\mathcal{\bar{S}}^{(1)}$ and $\mathcal{\bar{S}}^{(2)}$.  The instances are identical except that $\mathcal{\bar{S}}^{(1)}$ contains the system under test, while $\mathcal{\bar{S}}^{(2)}$ contains the baseline system.  In particular, they contain identical models of the environment with which the test systems interface.  The simulators are driven by the same pseudorandom seed input, which leads to the same sequence of disturbances being drawn in the simulator when the behaviors of the test and baseline systems match.  When the behavior of the two systems diverge, the seed automatically allows different disturbances to be drawn from each simulator following their diverging states.
We define a combined simulator that contains the two parallel simulators $\mathcal{\bar{S}}^{(1)}$ and $\mathcal{\bar{S}}^{(2)}$.  They are both driven by the same input seed $\bar{x}$.  Each simulator produces its own set of outputs, which include the transition likelihood $\rho$, an indicator of whether an event occurred $e$, and the miss distance $d$.  These variables are combined in a reward function, where a single reward is provided to the reinforcement learner.  Finally, the \ac{mctssa} algorithm chooses seeds to maximize the reward it receives.  The superscripts on the variables $\rho,e,d$ and $\tau$ indicate the associated simulator.

\subsection{Reward Function}
The reward function combines the output from the two individual simulators to produce a single reward for the \ac{mctssa} learner.  The primary objective of the reward function is to maximize the difference in outcomes of the simulators, driving the first simulator to a failure event while keeping the second simulator away from one.  The secondary objective is to maximize the path likelihoods of the two simulators to produce the most likely paths.  The \ac{dast} reward function is given by Equation~\ref{eq:dastreward}.  

\begin{alignat}{3}
	\label{eq:dastreward}
	R(\rho^{(1)},e^{(1)},d^{(1)},\tau^{(1)}, \rho^{(2)},e^{(2)},d^{(2)},\tau^{(2)})=
		& \quad R_E \cdot \mathds{1}{\{\tau^{(1)} \wedge e^{(1)}\}} \nonumber\\
		& -d^{(1)} \cdot \mathds{1}{\{\tau^{(1)} \wedge \neg e^{(1)}\}} \nonumber\\
		& -R_E \cdot \mathds{1}{\{\tau^{(2)} \wedge e^{(2)}\}} \nonumber\\
		& +d^{(2)} \cdot \mathds{1}{\{\tau^{(2)} \wedge \neg e^{(2)}\}} \nonumber\\
		& +(\log \rho^{(1)} + \log \rho^{(2)}) \cdot \mathds{1}{\{\neg \tau\}}
\end{alignat}

The \ac{dast} reward function extends the \ac{ast} reward function to the differential setting and has a similar structure:  
\begin{itemize}
\item The first term in Equation~\ref{eq:dastreward} gives a non-negative reward $R_E$ to the learner if the first simulator $\mathcal{\bar{S}}^{(1)}$ terminates in an event.  
\item If $\mathcal{\bar{S}}^{(1)}$ terminates and an event did not occur, then the second term penalizes the agent by giving the negative miss distance $-d^{(1)}$.  
\item The third term gives $-R_E$ if $\mathcal{\bar{S}}^{(2)}$ terminates in an event. 
\item The fourth term gives $d^{(2)}$ if $\mathcal{\bar{S}}^{(2)}$ terminates and an event did not occur.  
\item To maximize the likelihoods of the paths, the fifth term gives the sum of the log transition likelihoods of both simulators.
\end{itemize}
The third and fourth terms are the negations of the first and second terms, respectively, applied to the second simulator $\mathcal{\bar{S}}^{(2)}$.  
The terminal state of the simulators are treated as \emph{absorbing}.  That is, once a simulator enters a terminal state and collects the terminal reward, it stays there for all subsequent transitions and collects zero reward for these transitions.  The Boolean variables $\tau^{(1)}$ and $\tau^{(2)}$ indicate whether $\mathcal{\bar{S}}^{(1)}$ and $\mathcal{\bar{S}}^{(2)}$ have terminated, respectively.  The combined simulator terminates when both simulators have terminated, i.e., $\tau=\textsc{IsTerminal}(\mathcal{\bar{S}})=\tau^{(1)} \wedge \tau^{(2)}$, where $\tau$ (without superscript) indicates whether the combined simulator has terminated.

Due to the seed-action abstraction, we can optimize the reward function using the \ac{mctssa} algorithm described in \Cref{alg:ast_mcts} as used in \ac{ast}.  Because the algorithm is based on scalar rewards and pseudorandom seeds, the internal details of the simulator are abstracted from the reinforcement learner.  As a result, no modifications to the algorithm are necessary.

\section{Aircraft Collision Avoidance Application}
\label{sec:application}

Aircraft collision avoidance systems are mandated on all large transport and cargo aircraft in the United States and other countries around the world to help prevent mid-air collisions.  Their operation has played a crucial role in the exceptional level of safety in the national airspace \parencite{Kuchar2007}.  The \acf{tcas} is the current international standard for airborne collision avoidance systems and has been very successful at protecting aircraft from mid-air collisions.  However, studies have revealed fundamental limitations in \ac{tcas} that prevent it from operating effectively in the next-generation airspace where the number of aircraft is expected to increase significantly \parencite{Kuchar2007}.  To address the growing needs of the national airspace, the \acf{faa} has decided to create a new aircraft collision avoidance system.  The next-generation \acf{acasx} was created promising a number of improvements over \ac{tcas} including a reduction in collision risk while simultaneously reducing the number of unnecessary alerts \parencite{Kochenderfer2012next}.  This research work was performed on prototypes of \ac{acasx} while the system was still being developed and tested.  On September $20^{\text{th}}$, 2018, the RTCA\footnote{RTCA was formerly known as the Radio Technical Commission for Aeronautics, but is now known simply as RTCA.} accepted \ac{acasx} to replace \ac{tcas} as the next standard for airborne collision avoidance and the system is expected to be widely deployed in the near future.

\ac{acasx} has been shown to be much more operationally suitable than \ac{tcas} \parencite{FAA2018}.  \Cref{tab:acasxvstcas} is a comparison of \ac{acasx} and \ac{tcas} on several key operational metrics evaluated on a number of simulated aircraft encounter datasets.  The data is excerpted from the results of many studies performed at \acf{mitll} and presented at RTCA \parencite{FAA2018}.  \ac{acasx} shows significant improvements in overall safety and alert rates compared to \ac{tcas}.  The primary metric of safety is the probability of a \acf{nmac}, $P_{\text{NMAC}}$.  \Iac{nmac} is defined as two aircraft coming closer than 500 feet horizontally and 100 feet vertically.  On large encounter datasets, \ac{acasx} is shown to reduce the probability of \ac{nmac} by 17\% to 54\% compared to \ac{tcas}.  Alert metrics show significant improvements over \ac{tcas} as well, including the aggregate alert rate, which is the number of times the collision avoidance system alerts (counted by aircraft); and the 500 feet corrective alert rate, which is the number of alerts issued in encounters where the aircraft are flying level and vertically separated by exactly 500 feet.  \ac{acasx} also significantly improves metrics on generally undesirable scenarios such as the altitude crossing rate, which is the number of encounters where two aircraft cross in altitude; and the reversal rate, which is the number of times the collision avoidance system first advises the pilot to maneuver in one direction, then later advises the pilot to maneuver in the opposite direction.

\begin{table}[!htbp]
\small
\centering
\caption{A comparison of ACAS X and TCAS operational metrics on various encounter datasets. The data is excerpted from \parencite{FAA2018}.  ACAS X significantly improves overall safety, alert rates, and other operational metrics compared to TCAS.}
\label{tab:acasxvstcas}
\begin{tabular}{llcccc}
	\toprule
    Metric & Dataset & Number of & TCAS v7.1 & ACAS Xa & Improvement\\ 
    & & encounters & & 0.10.3 & over TCAS\\ 
    \midrule
    Safety ($P_{\text{NMAC}}$) & LLCEM & 5,956,128 & $2.179 \cdot 10^{-4}$ & $1.744 \cdot 10^{-4}$ & 19.96\% \\
    Safety ($P_{\text{NMAC}}$) & SAVAL & 75,173,906 & $4.361 \cdot 10^{-4}$ & $3.627 \cdot 10^{-4}$ & 16.83\% \\
    Safety ($P_{\text{NMAC}}$) & SA01 & 100,000 & $4.106 \cdot 10^{-2}$ & $1.873 \cdot 10^{-2}$ & 54.38\% \\
    Alert Rate (by aircraft)& TRAMS & 293,101 & 252,656 & 121,267 & 52.00\% \\
    500' Corrective Alert Rate & TRAMS & 175,184 & 14,912 & 9,919 & 33.48\% \\
    Altitude Crossing Rate & TRAMS & 293,101 & 3,196 & 1,582 & 50.50\% \\
    Reversal Rate & TRAMS & 293,101 & 1,029 & 556 & 45.97\% \\
	\bottomrule	
\end{tabular}
\end{table}

Studies have shown that the risk of \ac{nmac} is extremely small and, moreover, that \ac{acasx} reduces the risk even further than \ac{tcas} overall.  However, the risk of \ac{nmac} cannot be completely eliminated due to factors such as surveillance noise, pilot response delay, and the need for an acceptable alert rate \parencite{Kochenderfer2012next}.  Because \acp{nmac} are such important safety events, it is important to study and understand the rare circumstances under which they can still occur even if they are extremely unlikely.  Understanding the nature of the residual \ac{nmac} risk has been important for certification and informing the iterative development of the system.  


\subsection{ACAS X Operation}
There are several versions of \ac{acasx} under development.  This article considers a development (and not final) version of ACAS Xa, which uses active surveillance and is designed to be a direct replacement to \ac{tcas}.  Despite the internal logics of \ac{acasx} and \ac{tcas} being derived completely differently, the input and output interfaces of these two systems are identical.  
As a result, the following description of aircraft collision avoidance systems applies to both \ac{acasx} and \ac{tcas}.

Airborne collision avoidance systems monitor the airspace around an aircraft and issue alerts to the pilot if a conflict with another aircraft is detected.  These alerts, called \acp{ra}, instruct the pilot to maneuver the aircraft to  a certain target vertical velocity and maintain it.  The advisories are typically issued when the aircraft are within approximately 20--40 seconds to a potential collision.  \Cref{tab:acasxra} lists the possible primary \acp{ra}.  We use $\dot{z}_{\text{own}}$ to denote the current vertical velocity of own aircraft.  Own aircraft refers to the aircraft on which the collision avoidance system operates, whereas an intruding aircraft, or \emph{intruder}, is a nearby aircraft that poses a collision risk to own aircraft.  Intruders may or may not be equipped with their own collision avoidance systems.

\begin{table}[!htbp]
\small
\centering
\caption{Primary ACAS X advisories}
\label{tab:acasxra}
\begin{tabular}{llc}
	\toprule
    Abbreviation & Description & Rate to Maintain (ft/min)\\
    \midrule
    COC & clear of conflict & N/A\\
    DND & do not descend & $0$\\
    DNC & do not climb & $0$\\
    DND$x$ & do not descend at greater than $x$ ft/min& $\max(-x,\dot{z}_{\text{own}})$\\
    DNC$x$ & do not climb at greater than $x$ ft/min& $\min(x,\dot{z}_{\text{own}})$\\
    MAINTAIN & maintain current rate & $\dot{z}_{\text{own}}$\\
    DS1500 & descend at 1,500 ft/min & $-1,500$\\
    CL1500 & climb at 1,500 ft/min & $+1,500$\\
    DS2500 & descend at 2,500 ft/min & $-2,500$\\
    CL2500 & climb at 2,500 ft/min & $+2,500$\\
	\bottomrule	
\end{tabular}
\end{table}

The COC advisory stands for ``clear of conflict'' and is equivalent to no advisory. The pilot is free to choose how to control the aircraft.  The DND and DNC advisories stand for ``do not descend'' and ``do not climb'', respectively.  They restrict the pilot from flying in a certain vertical direction.  The DND$x$ and DNC$x$ advisories extend DND and DNC, respectively, to restrict the pilot from descending or climbing at a vertical rate greater than $x$ feet per minute.  The MAINTAIN advisory is preventative and instructs the pilot to maintain the current vertical rate of the aircraft.  The advisories DS1500 and CL1500 instruct the pilot to descend or climb at 1,500 feet per minute.  The pilot is expected to maneuver the aircraft at $\frac{1}{4}g$ acceleration until the target vertical rate is reached then maintain that vertical rate.  The DS2500 and CL2500 advisories instruct the pilot to descend or climb at an increased rate of 2,500 feet per minute.  These are strengthened advisories and demand a stronger response from the pilot.  For these strengthened \acp{ra}, the pilot is expected to maneuver at $\frac{1}{3}g$ acceleration until the target vertical rate is reached then maintain that vertical rate.  Strengthened \acp{ra} must follow a weaker \ac{ra} of the same vertical direction.  They cannot be issued directly.  For example, a CL1500 advisory must precede a CL2500 advisory.  Advisories issued by collision avoidance systems on different aircraft are not completely independent.  When \iac{ra} is issued, a coordination message is broadcast to other nearby equipped aircraft to prevent other collision avoidance systems from accidentally issuing \iac{ra} in the same vertical direction.  

\subsection{Experimental Setup}
\label{sec:setup}
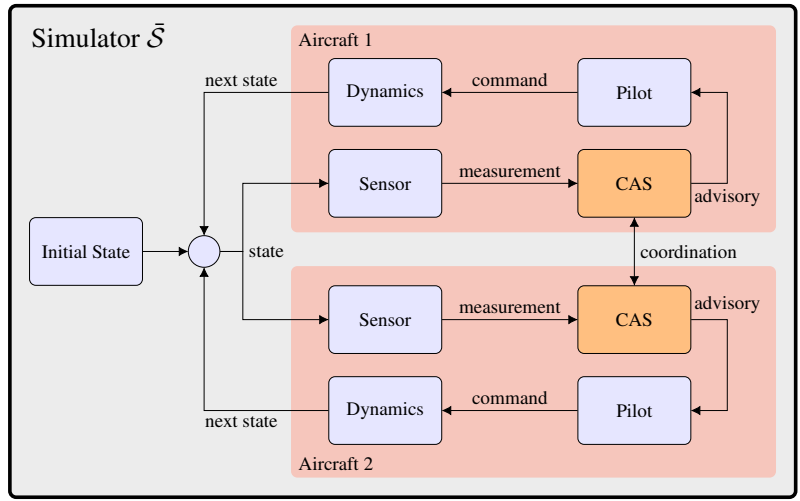
\begin{figure}[!htbp]
\centering
\resizebox{0.8\columnwidth}{!}
{
\begin{tikzpicture}[node distance=1.5cm, every node/.style={font=\large}, align=center]
	\node (initial) [componentstyle] {Initial State};
	\node (sum) [componentstyle,right=1cm of initial,circle,minimum size=0.7cm] {};
	\node (sensor1) [componentstyle,above right=0.5cm and 2.5cm of sum] {Sensor};
	\node (cas1) [casstyle,right=3cm of sensor1] {CAS};
	\node (pilot1) [componentstyle,above=0.5cm of cas1] {Pilot};
	\node (dynamics1) at (sensor1|-pilot1) [componentstyle] {Dynamics};
    
    \begin{pgfonlayer}{aircraftlayer}
        \node (aircraft1) [fit={($(dynamics1.north west)+(-7mm,6mm)$) ($(cas1.south east)+(1.75cm,-2mm)$)},aircraftstyle]{};
    	\node at ($(aircraft1.north west)+(1cm,-3.2mm)$) [] {Aircraft 1};
    \end{pgfonlayer}
    \draw[->] (initial) -- (sum);
    \draw[->] (sum.east) -- ++(0.5cm,0) node[right]{state} |- (sensor1.west);
    \draw[->] (sensor1) -- (cas1) node[midway,above]{measurement};
    \draw[->] (pilot1) -- (dynamics1) node[midway,above]{command};
    \draw[->] (dynamics1) -| (sum) node[pos=0.35,above]{next state};
    \draw[->] (cas1.east) -- ++(0.8cm,0) node[below]{advisory} |- (pilot1.east);
    
	\node (sensor2) [componentstyle,below right=0.5cm and 2.5cm of sum] {Sensor};
	\node (cas2) [casstyle,right=3cm of sensor2] {CAS};
	\node (pilot2) [componentstyle,below=0.5cm of cas2] {Pilot};
	\node (dynamics2) at (sensor2|-pilot2) [componentstyle] {Dynamics};
    
    \begin{pgfonlayer}{aircraftlayer}
        \node (aircraft2) [fit={($(sensor2.north west)+(-7mm,3mm)$) ($(pilot2.south east)+(1.75cm,-6mm)$)},aircraftstyle]{};
    	\node at ($(aircraft2.south west)+(1cm,3.2mm)$) [] {Aircraft 2};
    \end{pgfonlayer}
    \draw[->] (sum.east) -- ++(0.5cm,0) |- (sensor2.west);
    \draw[->] (sensor2) -- (cas2) node[midway,above]{measurement};
    \draw[->] (pilot2) -- (dynamics2) node[midway,above]{command};
    \draw[->] (dynamics2) -| (sum) node[pos=0.35,below]{next state};
    \draw[->] (cas2.east) -- ++(0.8cm,0) node[above]{advisory} |- (pilot2.east);
    
    \draw[<->] (cas1) -- (cas2) node[pos=0.5,right]{coordination};
    \begin{pgfonlayer}{simlayer}
        \path let \p1=(initial.west), \p2=(aircraft1.north east) in node (sim) [fit={($(\x1,\y2)+(-3mm,3mm)$) ($(aircraft2.south east)+(3mm,-3mm)$)},simstyle,line width=0.08cm]{};
    	\node at ($(sim.north west)+(2cm,-7mm)$) [font={\LARGE}] {Simulator $\mathcal{\bar{S}}$};
    \end{pgfonlayer}
    
\end{tikzpicture}
}
\caption{System diagram for pairwise encounters.  Two aircraft simulation loops are used.  Simulation models sensors, collision avoidance system, pilot response, aircraft dynamics, and interactions.}
\label{fig:acasxblock}
\end{figure}

We construct a seed-action simulator modeling an aircraft mid-air encounter.  We focus on encounters where all aircraft are equipped with identical collision avoidance systems. 
The overall architecture of the simulator for two aircraft is shown in \Cref{fig:acasxblock}.  
Simulation models capture the key aspects of the encounter, including the initial state, sensors, collision avoidance system, pilot response, and aircraft dynamics.  

\subsubsection{Simulation Model}
\label{sec:simulator}

\paragraph{\textbf{Initial State.}}  The initial state of the encounter includes initial positions, velocities, and headings of the aircraft.  The initial state is drawn from a distribution that gives realistic initial configurations of aircraft that are likely to lead to \ac{nmac}.  Once the initial state is sampled, it is fixed for the duration of the search.
In our experiments, pairwise (two-aircraft) encounters are initialized using the \acf{llcem} \parencite{Kochenderfer2008cor,Kochenderfer2010jgcd}.  \ac{llcem} is a statistical model learned from a large body of radar data of the entire national airspace.  We follow the encounter generation procedure described in the paper \parencite{Kochenderfer2008cor}.
Multi-threat (three-aircraft) encounters use the \emph{star model}, which initializes aircraft on a circle heading towards the origin, spaced apart at equal angles.  Initial airspeed, altitude, and vertical rate are sampled from a uniform distribution over a prespecified range.  The horizontal distance from the origin is set such that without intervention, the aircraft intersect at approximately 40 seconds into the encounter. 

\paragraph{\textbf{Sensor Model.}}  The sensor model captures how the collision avoidance system perceives the world.  We assume active, beacon-based radar capability with no noise.  For own aircraft, the sensor measures the vertical rate, barometric altitude, heading, and height above ground.  For each intruding aircraft, the sensor measures slant range (relative distance to intruder), bearing (relative horizontal angle to intruder), and relative altitude.

\paragraph{\textbf{Collision Avoidance System.}}  The collision avoidance system is the system under test.  We use a prototype of \ac{acasx} in the form of a binary library obtained from the \ac{faa}.  The binary has a minimal interface that allows initializing and stepping the state forward in time.  The system maintains internal state, but does not expose it.  The primary output of the \ac{acasx} system is the \ac{ra}. 
\ac{acasx} has a coordination mechanism to ensure that issued \acp{ra} from different aircraft are compatible with one another, i.e., that two aircraft are not instructed to maneuver in the same vertical direction.  The messages are communicated to all nearby aircraft through coordination messages.  
Our differential studies compare \ac{acasx} to \ac{tcas}.  Our implementation of \ac{tcas} is also a binary library obtained from the \ac{faa}.  Both binaries have identical input and output interfaces making them interchangeable in the simulator.

\paragraph{\textbf{Pilot Model.}}  The pilot model consists of a model for the pilot's intent and a model for the pilot's response to \iac{ra}.  The pilot's intent is how the pilot would fly the aircraft if there were no \acp{ra}.  To model intended commands, we use the pilot command model in \ac{llcem}, which gives a realistic stochastic model of aircraft commands in the airspace \parencite{Kochenderfer2008cor}.  The pilot response model defines how pilots respond to an issued \ac{ra}.  Pilots are assumed to respond deterministically to \iac{ra} with perfect compliance after a fixed delay \parencite{ICAO2007}.  Pilots respond to initial \acp{ra} with a five-second delay and subsequent \acp{ra} (i.e., strengthenings and reversals) with a three-second delay.  During the initial delay period, the pilot continues to fly the intended trajectory.  During response delays from subsequent \acp{ra}, the pilot continues responding to the previous \ac{ra}.  Multiple \acp{ra} issued successively are queued so that both their order and timing are preserved.  In the case where a subsequent \ac{ra} is issued within 2 seconds or less of an initial \ac{ra}, the timing of the subsequent \ac{ra} is used and the initial \ac{ra} is skipped.  The pilot command includes commanded airspeed acceleration, commanded vertical rate, and commanded turn rate. 

\paragraph{\textbf{Aircraft Dynamics Model.}}  The aircraft dynamics model determines how the state of the aircraft propagates with time.  The aircraft state includes the airspeed, position north, position east, altitude, roll angle, pitch angle, and heading angle.  The aircraft state is propagated forward at \SI{1}{\hertz} using forward Euler integration.

In our experiments, the commands of the pilots when not responding to an advisory are stochastic and are being optimized by the algorithm.  When no \ac{ra} is active, the pilot commands follow the stochastic dynamic model of \ac{llcem}.  When \iac{ra} is active, the vertical component of the pilot's command follows the pilot response model, while the other components follow \ac{llcem}.  Other simulation components are deterministic in our experiments.  However, in the future, we may consider stochastic models for sensors, aircraft dynamics, and pilot response.

\subsubsection{Reward Function}
We use the reward function defined in Equation~\ref{eq:poastreward} for optimization.  The event space $E$ is defined to be \iac{nmac}, which occurs when two aircraft are closer than 100 feet vertically and 500 feet horizontally.  We define the miss distance $d$ to be the distance of closest approach, which is the Euclidean distance of the aircraft at their closest point in the encounter.  The distance of closest approach is a good metric because it is monotonically decreasing as trajectories get closer and reaches a minimum at \iac{nmac}.  
 
\subsection{Stress Testing Single-Threat Encounters}
\label{sec:singlethreat}
We apply \ac{ast} to analyze \acp{nmac} in encounters involving two aircraft.  We searched 100 encounters initialized using samples from \ac{llcem}.  The configuration is shown in \Cref{tab:singlethreatconfig}.  Of the 100 encounters searched, 18 encounters contained \iac{nmac}, yielding an empirical find rate of 18\%.  When the optimization algorithm completes, it returns the path with the highest reward regardless of whether the path contains \iac{nmac}.  When the returned path does not contain \iac{nmac}, it is uncertain whether \iac{nmac} exists and the algorithm was unable to find it, or whether no \ac{nmac} is reachable given the initial state of the encounter.  
We manually cluster the \ac{nmac} encounters and present our findings.  

\begin{table}[!htbp]
\small
\centering
\caption{Single-threat configuration}
\label{tab:singlethreatconfig}
\begin{tabular}{ll}
	\toprule
    Simulation & \\ \midrule
    number of aircraft & 2\\
	initialization & \acs{llcem} \\
    sensors & active, beacon-based, noiseless\\
    collision avoidance system & ACAS Xa Run 13 libcas 0.8.5\\
	pilot response model & deterministic 5s--3s\\ \midrule
    MCTS & \\ \midrule
    maximum steps & 50\\
    iterations & 2000\\
    exploration constant & 100.0\\
    $k$ & 0.5\\
    $\alpha$ & 0.85\\
	\bottomrule	
\end{tabular}
\end{table}

\begin{figure}[!htbp]
\centering
\resizebox{0.85\columnwidth}{!}
{
\input{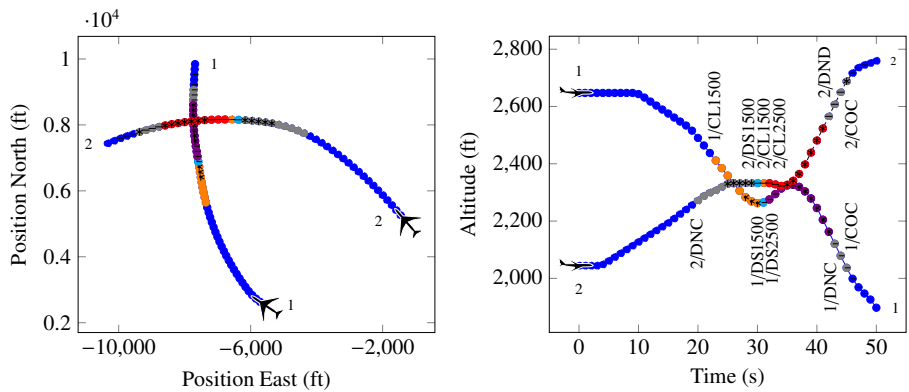}
}
\caption{NMAC encounter where the aircraft cross in altitude during pilot delay.  The aircraft cross in altitude after aircraft 1's \ac{ra} has been issued, but before they start to respond. The aircraft starts to climb from below the intruder and the encounter ends in \iac{nmac} at 36 seconds.}
\label{fig:pairwise4}
\end{figure}

\paragraph{\textbf{Crossing Time.}}  We observed a number of \acp{nmac} resulting from well-timed vertical maneuvers.  In particular, several encounters included aircraft crossing in altitude during the pilot response delay to an initial \ac{ra}.  \Cref{fig:pairwise4} shows one such encounter that eventually ends in \iac{nmac} at 36 seconds into the encounter.  
The probability density of the encounter evaluated under \ac{llcem} is $5.3 \cdot 10^{-18}$.  
This measure can be used as an unnormalized measure of likelihood of occurrence.  In this encounter, the aircraft cross in altitude during pilot 1's response delay.  The crossing leads to aircraft 1 starting the climb from below aircraft 2.  The subsequent reversal later in the encounter is unable to resolve the conflict due to the pilot response delay.

\ac{nmac} encounter plots contain horizontal tracks (top-down view) and vertical profile (altitude versus time).  Aircraft numbers are indicated at the start and end of each trajectory.  \ac{ra} codes are labeled at the time of occurrence.  Marker colors indicate the \ac{ra} issued: blue for COC, orange for CL1500, red for CL2500, cyan for DS1500, purple for DS2500, and gray for DNC/DND/MAINTAIN/MULTITHREAT.  We prepend the aircraft number followed by a slash to each \ac{ra} code for readability since the aircraft may cross multiple times during the encounter.  Symbols inside the markers indicate the state of the pilot response: no symbol for not responding, dash for responding to previous \ac{ra}, and asterisk for responding to current \ac{ra}.

\paragraph{\textbf{Maneuvering Against RA.}}  Our analysis revealed a number of \ac{nmac} encounters where the pilot initially maneuvers against the issued \ac{ra} before complying.  That is, after the pilot receives the \ac{ra}, they maneuver the aircraft in the opposite direction of what is instructed by the \ac{ra} for the duration of the pilot response delay before subsequently complying and reversing direction.  Pilots do not normally maneuver in this manner and so this scenario is very operationally rare.  Even so, \ac{acasx} does seem to be able to resolve the majority of these initially disobeying cases.  In most cases, the maneuvering against the \ac{ra} must be very aggressive to result in \iac{nmac}.  

\begin{figure}[!htbp]
\centering
\centering
\resizebox{0.85\columnwidth}{!}
{
\input{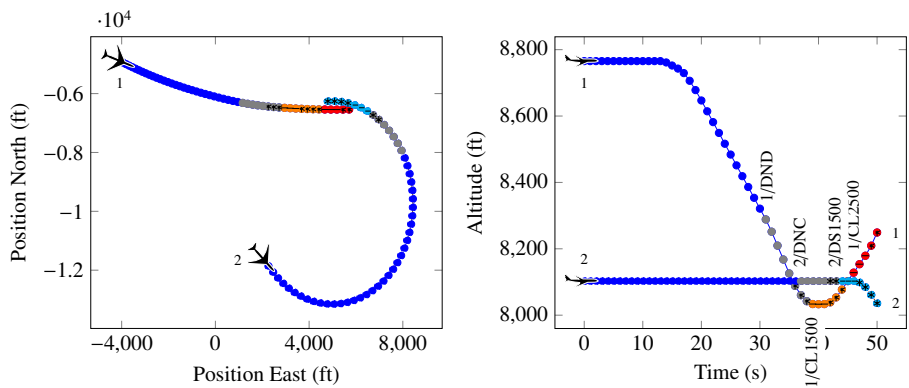}
}
\caption{NMAC encounter where one aircraft is turning at high rate while the other is descending at high vertical rate.  The encounter ends in \iac{nmac}, but the combination is operationally very rare.}
\label{fig:pairwise13}
\end{figure}

\paragraph{\textbf{High Turn and Vertical Rates.}}  Turns at high rates quickly shorten the time to closest approach. \ac{acasx} does not have full state information about its intruder and must estimate it by tracking relative distance, relative angle, and the intruder altitude.  \Cref{fig:pairwise13} shows an example of an encounter that has similar crossing behavior as \Cref{fig:pairwise4} but exacerbated by the high turn rate of aircraft 2 (approximately 1.5 times the standard turn rate).  In this scenario, the aircraft become almost head-on at the time of closest approach and a reversal is not attempted. \Iac{nmac} with a probability density of $6.5 \cdot 10^{-17}$ occurs at 48 seconds into the encounter.  Aircraft 1 is also coincidently descending at a high vertical rate.  The combination makes this encounter operationally very unlikely.

\paragraph{\textbf{Sudden Aggressive Maneuvering.}}  Sudden maneuvers can lead to \acp{nmac} if they are sufficiently aggressive.  In particular, we observed some \ac{nmac} encounters where two aircraft are approaching one another separated in altitude and flying level, then one aircraft suddenly accelerates vertically towards the other aircraft as they are about to pass.  Under these circumstances, given the pilot response delays and dynamic limits of the aircraft, there is insufficient time and distance remaining for the collision avoidance system to resolve the conflict.  Pilots do not normally fly so aggressively in operation, so this case is extremely unlikely. 
In fact, such maneuvers are even more rare than our model predicts.  \ac{acasx} issues \acp{ta} to alert pilots to nearby traffic, so that pilots are made aware of intruding aircraft well before the initial \ac{ra}.  These advance warnings increase the pilot's situational awareness and reduce blunders like these.  Our simulator does not model the effect of such \acp{ta}, however.
For a designer of \ac{acasx}, one course of action would be to tune \ac{acasx} to intervene preemptively in such scenarios.  While this reduces the risk of possible sudden behavior, it also increases the alert rate of the system.  Ultimately, the \ac{acasx} designer must assess the probabilities of various scenarios and find the delicate balance between risk of collision and issuing too many advisories.  Since these scenarios are extremely rare, we must trade off accordingly.

\paragraph{\textbf{Combined Factors.}}  In our experiments, it is rare for \iac{nmac} to be attributable to a single cause.  More commonly, a combination of factors contribute to the \ac{nmac}.  \Cref{fig:pairwise4} shows an example of an encounter where multiple factors contribute to the \ac{nmac}.  In \Cref{fig:pairwise4}, crossing time played a crucial role in the \ac{nmac}.  However, there are a number of other factors that are important as well.  The behavior where the aircraft turn horizontally towards each other is significant as it reduces the time to closest approach.  The two vertical maneuvers of aircraft 1 before receiving \iac{ra} are also important.  Similar observations can be made for many of the other \ac{nmac} encounters found.

\subsection{Stress Testing Multi-Threat Encounters}
\label{sec:multithreat}
We applied \ac{ast} to analyze \acp{nmac} in three-aircraft encounters.  We searched 100 encounters initialized using samples from the star model.  The configuration is shown in \Cref{tab:multithreatconfig}.  We found 25 \acp{nmac} out of 100 encounters searched, yielding a find rate of 25\%.

\begin{table}[!htbp]
\small
\centering
\caption{Multi-threat configuration}
\label{tab:multithreatconfig}
\begin{tabular}{ll}
	\toprule
    Simulation & \\ \midrule
    number of aircraft & 3\\
	initialization & star model\\
    sensors & active, beacon-based, noiseless\\
    collision avoidance system & ACAS Xa Run 13 libcas 0.8.5\\
	pilot response model & deterministic 5s--3s\\ \midrule
    MCTS & \\ \midrule
    maximum steps & 50\\
	iterations & 1000\\
    exploration constant & 100.0\\
    $k$ & 0.5\\
    $\alpha$ & 0.85\\
	\bottomrule	
\end{tabular}
\end{table}

\paragraph{\textbf{Limited Maneuverable Space.}}  In general, multi-threat encounters are more challenging to resolve than pairwise encounters because there is less open space for the aircraft to maneuver.  \Cref{fig:multithreat3} shows an example of \iac{nmac} encounter where aircraft 1 (the aircraft in the middle altitude between 10 and 36 seconds) needs to simultaneously avoid an aircraft below and a vertically closing aircraft from above.  \Iac{nmac} with a probability density of $1.0 \cdot 10^{-16}$ occurs at 39 seconds into the encounter.  Aircraft 2's downward maneuver greatly reduces the maneuverable airspace of aircraft 1.  
These encounters are undoubtedly extremely challenging for a collision avoidance system and it is unclear whether any satisfactory resolution exists.  Nevertheless, we gain insight by observing how the collision avoidance system behaves under such extremely rare circumstances.  

\begin{figure}[!htbp]
\centering\resizebox{0.85\columnwidth}{!}
{
\input{multithreat41.tikz}
}
\caption{NMAC encounter where the aircraft have limited maneuverable airspace.  Aircraft 1 must avoid aircraft 2 during maneuver away from aircraft 3.  The encounter ends in \ac{nmac} at 39 seconds.}
\label{fig:multithreat3}
\end{figure}

\paragraph{\textbf{Pairwise Coordination in Multi-Threat.}}  Our algorithm discovered a number of \ac{nmac} encounters where all aircraft are issued a MULTITHREAT (MTE) \ac{ra} and asked to follow an identical climb rate.  Complying with the \ac{ra} results in the aircraft closing horizontally without gaining vertical separation.  \Cref{fig:multithreat55} shows an example of such an encounter where \iac{nmac} occurs with probability density $5.8 \cdot 10^{-7}$ at 38 seconds into the encounter.
In discussing these results with the \ac{acasx} development team, we learned that this behavior is a known issue that can arise when performing multi-aircraft coordination using a pairwise coordination mechanism.  The pairwise coordination messages in essence determine which aircraft will climb and which will descend in an encounter. Since coordination messaging occurs pairwise, under rare circumstances it is possible for each aircraft to receive conflicting coordination messages from the other aircraft in the scenario. In nominal encounters, the aircraft that receives conflicting coordination messages from two aircraft remains level and lets the other aircraft climb or descend around it.  However, in these encounters, all three aircraft receive conflicting coordination messages. 
Although very rare, this is an important case that is being addressed by both \ac{tcas} and \ac{acasx} development teams.


\begin{figure}[!htbp]
\centering
\resizebox{0.85\columnwidth}{!}
{
\input{multithreat7.tikz}
}
\caption{NMAC encounter where all aircraft receive pairwise conflicting coordination messages. The encounter ends in \iac{nmac} at 38 seconds.  This is a very rare but important case that the \ac{acasx} team is addressing.} 
\label{fig:multithreat55}
\end{figure}

\paragraph{\textbf{Pairwise Phenomena.}}  Phenomena that appear in pairwise encounters also appear in multi-threat encounters. The presence of the third aircraft typically exacerbates the encounter.   In our multi-threat analysis, we noted similar phenomena related to crossing time, maneuvering against \ac{ra}, and sudden aggressive maneuvering as discussed previously.  We did not observe any cases related to high turn rates in the multi-threat setting due to our use of the star model.

\clearpage
\subsection{Stress Testing ACAS X Relative to TCAS Baseline}
\label{sec:dast_results1}
We apply \ac{dast} to perform differential stress testing of \ac{acasx} relative to a \ac{tcas} baseline.  We seek to find \ac{nmac} encounters that occur in \ac{acasx} but not in \ac{tcas}.  The search is extremely difficult. Not only are both \ac{acasx} and \ac{tcas} systems extremely safe, which means that \acp{nmac} are extremely rare, but also \ac{acasx} is a much safer system than \ac{tcas} overall, which makes cases where \ac{acasx} has \ac{nmac} but \ac{tcas} does not extremely rare.  We seek to find those extremely rare corner cases. 

We searched 2700 pairwise encounters initialized with samples from \ac{llcem}.  The configuration is shown in Table~\ref{tab:dastconfig}. The top 10 highest reward paths were returned for each encounter initialization producing a total of 27,000 paths.  Of these paths, a total of 28 contained \acp{nmac}, which originated from 10 encounter initializations.  We analyzed the scenarios and confirmed that all were operationally rare scenarios.
We present examples of \acp{nmac} found by the algorithm and discuss their properties.

\begin{table}[!htbp]
\small
\centering
\caption{Differential stress testing configuration}
\label{tab:dastconfig}
\begin{tabular}{ll}
	\toprule
    Simulation & \\ \midrule
    number of simulators & 2\\
    number of aircraft per simulator & 2\\
	initialization & LLCEM\\
    sensors & active, beacon-based, noiseless\\
    collision avoidance system (test) & ACAS Xa Run 15 libcas 0.10.3\\
    collision avoidance system (baseline) & TCAS II v7.1\\
	pilot response model & deterministic 5s--3s\\ \midrule
    MCTS & \\ \midrule
    maximum steps & 50\\
	iterations & 3000\\
    exploration constant & 100.0\\
    $k$ & 0.5\\
    $\alpha$ & 0.85\\
	\bottomrule	
\end{tabular}
\end{table}

\begin{figure}[!htbp]
\centering\resizebox{\columnwidth}{!}
{
\input{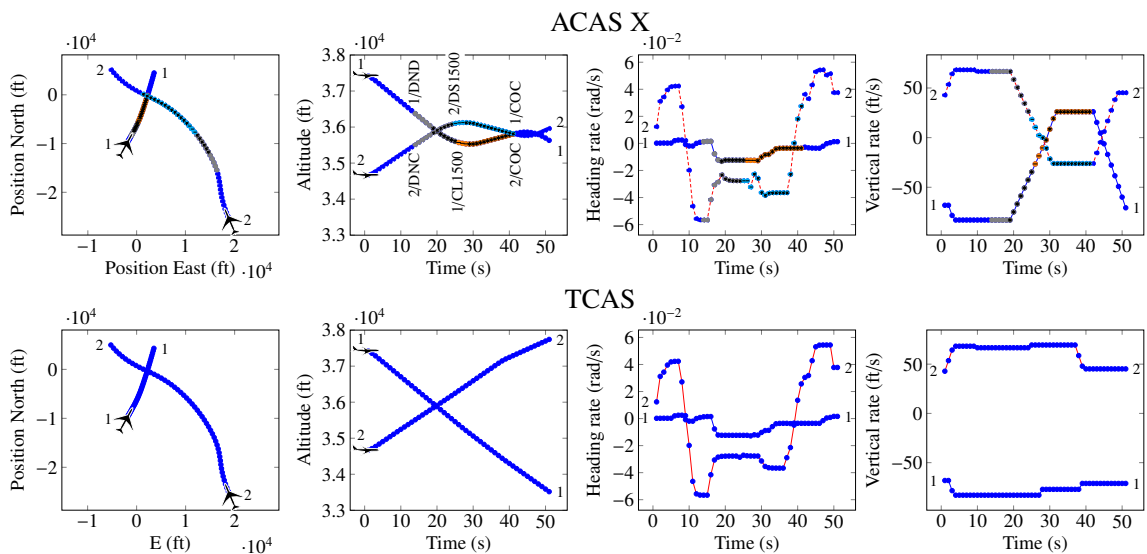}
}
\caption{An encounter where ACAS X issues an RA but TCAS does not.  The aircraft are initially traveling at high vertical rates and aircraft 2 also performs some horizontal maneuvering. \Iac{nmac} occurs in the \ac{acasx} simulation at 40 seconds.}
\label{fig:traj430_k1}
\end{figure}

\paragraph{\textbf{ACAS X Issues RA, but TCAS Does Not.}} We found some \ac{nmac} cases where \ac{acasx} issued \iac{ra} but \ac{tcas} did not.  An example is shown in \Cref{fig:traj430_k1} where \iac{nmac} occurs at 40 seconds in the \ac{acasx} simulation but no \ac{nmac} occurs in the \ac{tcas} simulation.  
The encounter occurs with a probability density of $1.7 \cdot 10^{-19}$.  
No \ac{ra} was issued in the \ac{tcas} simulation.  In the example, both aircraft are traveling at a high absolute vertical rate exceeding 70 feet per second toward each other.  Both aircraft receive \iac{ra} to level-off but the aircraft cannot respond in time due to the high vertical rates and the pilot response delay.  The aircraft proceed to cross in altitude.  After crossing, \ac{acasx} increases the strength of the advisory in the same direction as the previous \ac{ra}.  Since the aircraft have crossed in altitude, responding to the \ac{ra} results in a loss of vertical separation and \iac{nmac}.  

High vertical rates are known to make conflict resolution more difficult, especially for vertical-only collision avoidance systems like \ac{tcas} and this version of \ac{acasx}.  
Aircraft with high vertical rates take longer to reverse direction vertically and more vertical distance is traveled during the pilot's response delay.  As a result, advisories take longer to take full effect.  Moreover, in cases where both aircraft are maneuvering in the same vertical direction, an aircraft may lose the ability to ``outrun'' the other aircraft.  For example, even if a maximal climb advisory is issued, it may not be sufficient for the aircraft to stay above a second aircraft climbing at an even higher rate.  Another interesting feature of this encounter is the horizontal behavior.  Aircraft 2 is initially turning away from the other aircraft before turning towards it at 8 seconds into the encounter.  The initial \ac{ra} is issued shortly after that maneuver. Large rapid changes in turn rate around the time of \iac{ra} can make it difficult for a collision avoidance system to accurately estimate the time to horizontal intersection.   

Overall, \ac{acasx} is much safer and more operationally suitable than \ac{tcas}.  However, there are some trade-offs between the two systems as highlighted by our methods.  \ac{acasx}'s late altering characteristic, which reduces the number of unnecessary alerts, can sometimes hurt encounters with higher vertical rates, such as seen in this example.  In deciding the trade-off, a designer must weigh the relative likelihood of these encounters versus the effect on other more frequently observed trajectories. 

\begin{figure}[!hbtp]
\centering\resizebox{1.0\columnwidth}{!}
{
\input{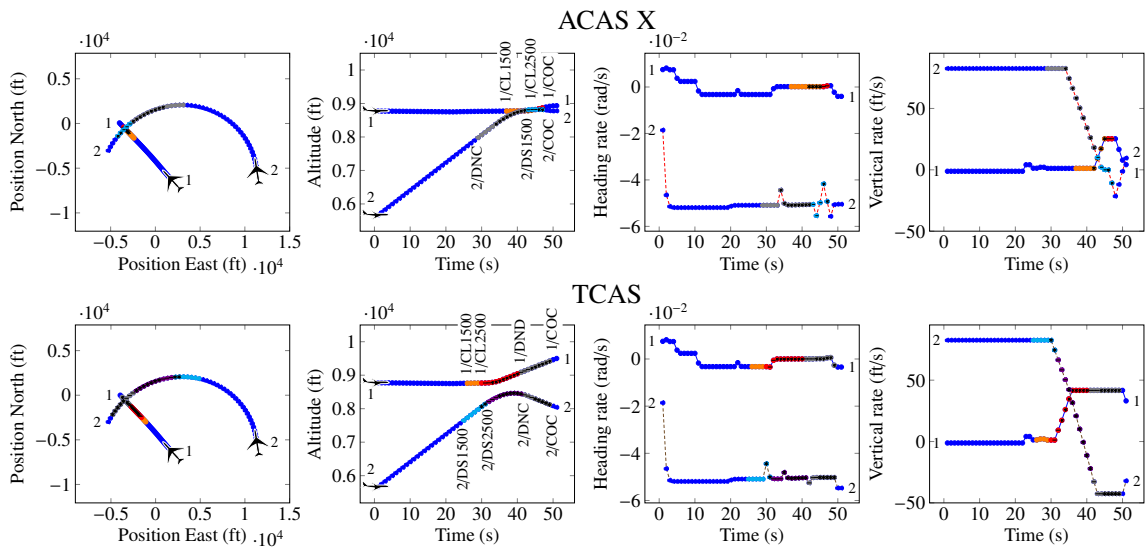}
}
\caption{An NMAC encounter where one aircraft turns while simultaneously climbing very rapidly. \Iac{nmac} occurs in the \ac{acasx} simulation at 45 seconds whereas no \ac{nmac} occurs in the \ac{tcas} simulation.}
\label{fig:traj991_k1}
\end{figure}

\paragraph{\textbf{Simultaneous Horizontal and Vertical Maneuvering.}} Many of the \ac{nmac} encounters found involve an aircraft turning while simultaneously climbing or descending very rapidly.  \Cref{fig:traj991_k1} shows an example where \iac{nmac} occurs at 45 seconds in the \ac{acasx} simulation but no \ac{nmac} occurs in the \ac{tcas} simulation.  
The encounter occurs with probability density $2.7 \cdot 10^{-4}$.
In this example,  aircraft 1 is flying generally straight and level while aircraft 2 is simultaneously turning and climbing at a vertical rate exceeding 80 feet per second.  Aircraft 2 receives \iac{ra} to level-off but is unable to maneuver in time before losing vertical separation resulting in \iac{nmac}.  In this example, \ac{tcas} is able to resolve the conflict by issuing \acp{ra} to both aircraft earlier in the encounter than \ac{acasx}.  

Horizontally, aircraft 2's turn quickly shortens the time to closest approach between the two aircraft.  Vertically, aircraft 2 receives a level-off advisory but the high climb rate and pilot response delay limit how quickly the aircraft can be brought to compliance.  Scenarios that involve turning and simultaneously climbing or descending at high rate are operationally very rare.

\begin{figure}[!hbtp]
\centering\resizebox{1.0\columnwidth}{!}
{
\input{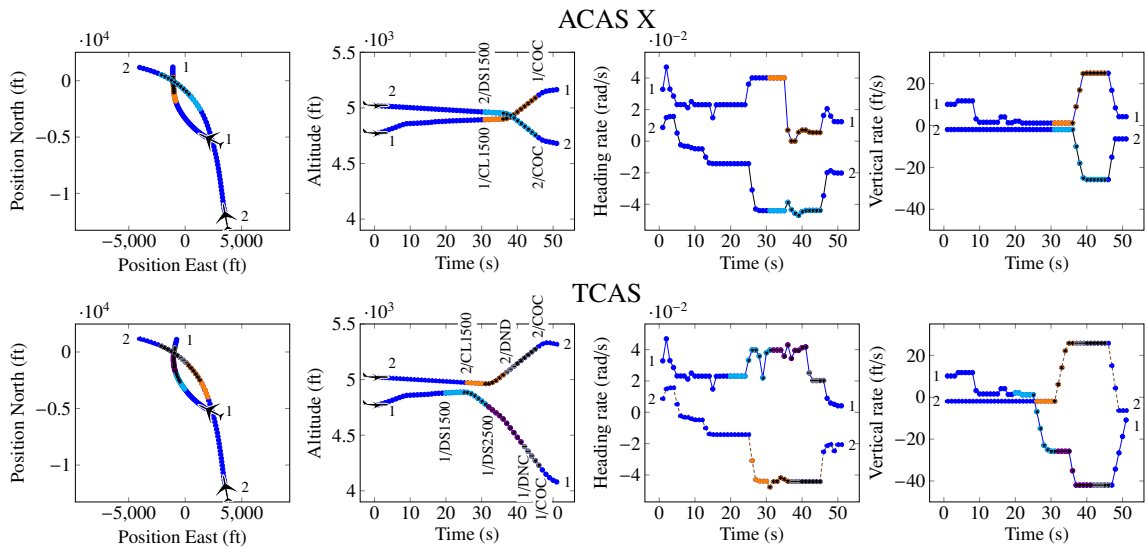}
}
\caption{An NMAC encounter where the aircraft maneuver horizontally only.  The aircraft start in an unlikely initial configuration.  \Iac{nmac} occurs at 40 seconds in the \ac{acasx} simulation but not in the \ac{tcas} simulation.}
\label{fig:traj5572_k1}
\end{figure}

\paragraph{\textbf{Horizontal Maneuvering.}} In some very rare cases, \acp{nmac} can also result from horizontal maneuvering alone.  An example is shown in \Cref{fig:traj5572_k1} where \iac{nmac} occurs at 40 seconds in the \ac{acasx} simulation but no \ac{nmac} occurs in the \ac{tcas} simulation.
The encounter occurs with probability density $4.3 \cdot 10^{-11}$.
The aircraft are initially headed away from each other but they are also turning towards each other.  At 25 seconds into the encounter, the aircraft turn more tightly towards each other, rapidly reducing the time to closest approach.  Crossing advisories are issued to the aircraft 9 seconds prior to \ac{nmac}.  However, there is not enough time remaining to cross safely and \iac{nmac} occurs.  
In this example, \ac{tcas} is able to resolve the conflict by issuing \acp{ra} to both aircraft earlier in the encounter.
However, it is unclear how the aircraft got to their initial positions in the encounter and whether this initial position is generally reachable.

\FloatBarrier
\subsection{Stress Testing TCAS Relative to ACAS X Baseline}
\label{sec:dast_results2}

For comparison, we perform \ac{dast} experiments where we consider \ac{tcas} as the system under test and \ac{acasx} as the baseline, which is the inverse of the experiment in \Cref{sec:dast_results1}.  We searched 2700 pairwise encounters with samples from \ac{llcem}.  We use the same configuration as in Table~\ref{tab:dastconfig} with the exception that the test and baseline systems are reversed.  The top 10 highest reward paths are returned for each encounter initialization producing a total of 27,000 paths as before.  Of these paths, a total of 39 contained \acp{nmac}, which represents an increase of 39.3\%.  It is reassuring that this result, where \ac{dast} found 39.3\% more \acp{nmac} with \ac{tcas} compared to \ac{acasx}, is qualitatively similar to the results from other studies which use direct Monte Carlo sampling and also show a safety benefit of \ac{acasx} compared to \ac{tcas} \parencite{Holland2013}.  The \acp{nmac} originate from a broader set of encounter initializations as well (22 versus 10 previously). 
We present examples of \acp{nmac} found by the algorithm and discuss their properties.

\begin{figure}[!hbtp]
\centering\resizebox{1.0\columnwidth}{!}
{
\input{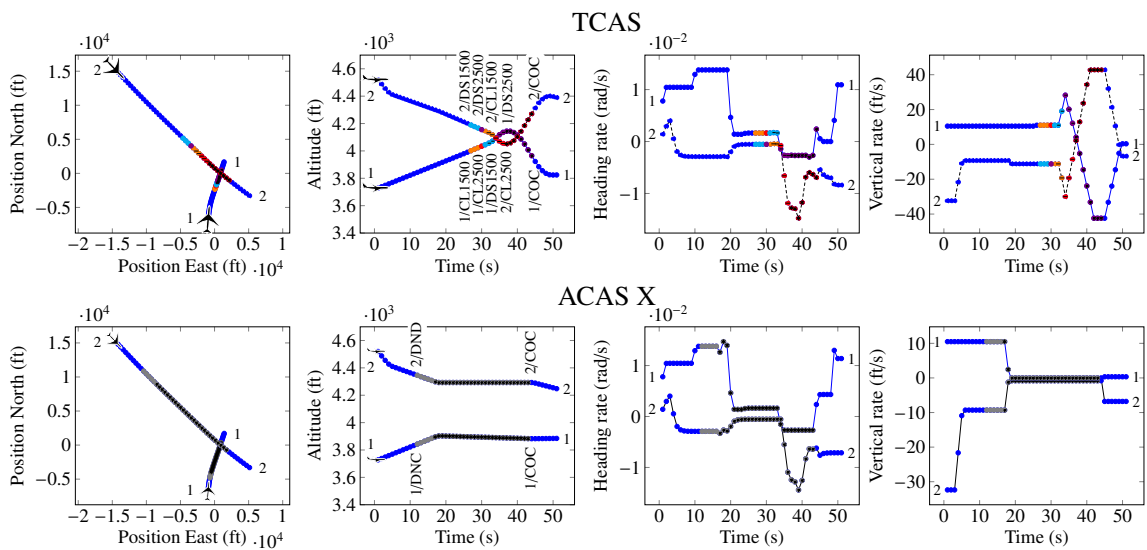}
}
\caption{An NMAC encounter where a crossing \ac{ra} is reversed prior to altitude crossing.  \Iac{nmac} occurs at 41 seconds in the \ac{tcas} simulation but not in the \ac{acasx} simulation.}
\label{fig:dast2_1244_k1}
\end{figure}

\paragraph{\textbf{Crossing RA Followed by a Reversal.}} An example is shown in \Cref{fig:dast2_1244_k1} where \iac{nmac} occurs at 41 seconds in the \ac{tcas} simulation  but no \ac{nmac} occurs in the \ac{acasx} simulation. The encounter occurs with a probability density of $6.6 \cdot 10^{-12}$.  In the encounter, the aircraft converge vertically while they approach in a perpendicular configuration horizontally.  \ac{tcas} issues crossing initial advisories to the aircraft, but then reverses the \acp{ra} shortly before the aircraft cross in altitude.  The reversal leads to the aircraft reversing vertical direction after they have crossed in altitude, reducing their vertical separation, and eventually resulting in \iac{nmac}.  In contrast, \ac{acasx} issues preventative advisories to the aircraft early in the encounter and keeps the aircraft vertically separated throughout the encounter. 

\newpage
The crossing \ac{tcas} advisory combined with a reversal shortly afterwards suggests that the encounter may be operating near a decision boundary and \ac{tcas} may be having trouble deciding whether the aircraft should cross in altitude.  In this case, it appears the initial crossing \ac{ra} may have successfully resolved the conflict had it been maintained, and it is the reversal that complicated the resolution.  The lateness of the \ac{tcas} initial \ac{ra} likely played a major role in the \ac{nmac}, leaving a very difficult decision for the direction selection of the \ac{ra}.  \ac{acasx} gives a desirable outcome in this case, deciding early in the encounter that the aircraft should not cross in altitude.  

\begin{figure}[!hbtp]
\centering\resizebox{1.0\columnwidth}{!}
{
\input{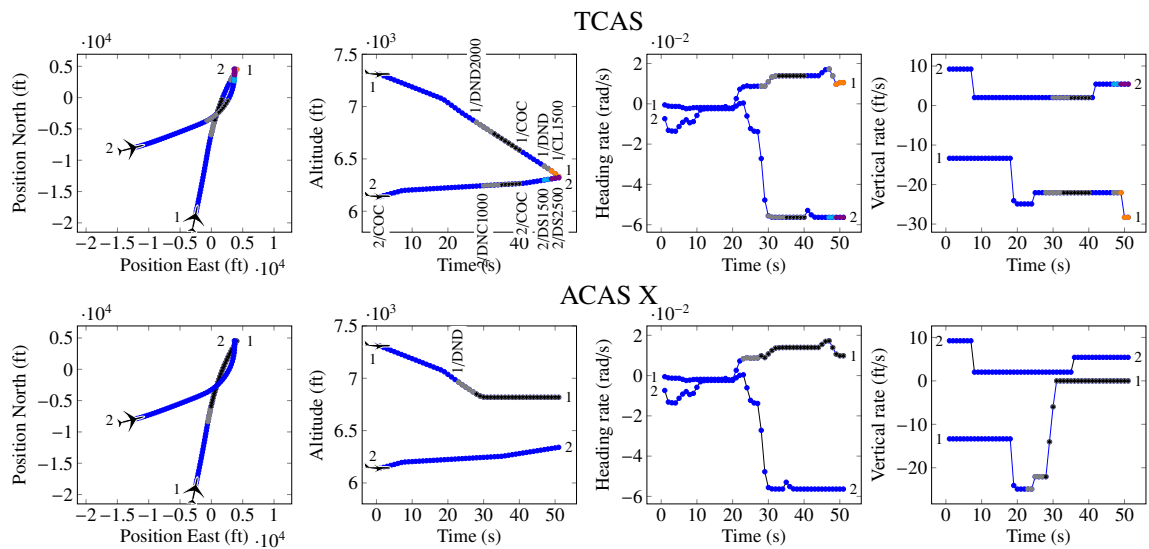}
}
\caption{An NMAC encounter with a double horizontal crossing and a split advisory.  \Iac{nmac} occurs at 50 seconds in the \ac{tcas} simulation but not in the \ac{acasx} simulation.}  
\label{fig:dast2_1190_k1}
\end{figure}

\paragraph{\textbf{Double Horizontal Crossing.}} An example is shown in \Cref{fig:dast2_1190_k1} where \iac{nmac} occurs at 50 seconds in the \ac{tcas} simulation, but no \ac{nmac} occurs in the \ac{acasx} simulation.  The encounter occurs with a probability density of $2.2 \cdot 10^{-14}$.   In the encounter, the aircraft turn as they converge both horizontally and vertically, eventually resulting in a horizontal double crossing.  \ac{tcas} issues preventative DND2000 and DNC1000 advisories to the aircraft, respectively, predicting that they will cross safely, then further clears the advisories after the first horizontal crossing.  However, after the first crossing, the aircraft continue to turn toward each other into a second horizontal crossing.  \ac{tcas} issues a second sequence of advisories, but it is too late and \iac{nmac} occurs.  Sequences of \acp{ra} separated by COC are called \textit{split advisories} and are undesirable because they may cause confusion or doubt in the pilots.  Another contributor to the difficulty of the encounter is aircraft 2 increasing its turn rate mid-encounter, which significantly reduces time to collision. 
The \ac{acasx} system handles the encounter much more desirably, choosing a DND advisory to maintain vertical separation as the aircraft cross horizontally.  

\begin{figure}[!htbp]
\centering\resizebox{1.0\columnwidth}{!}
{
\input{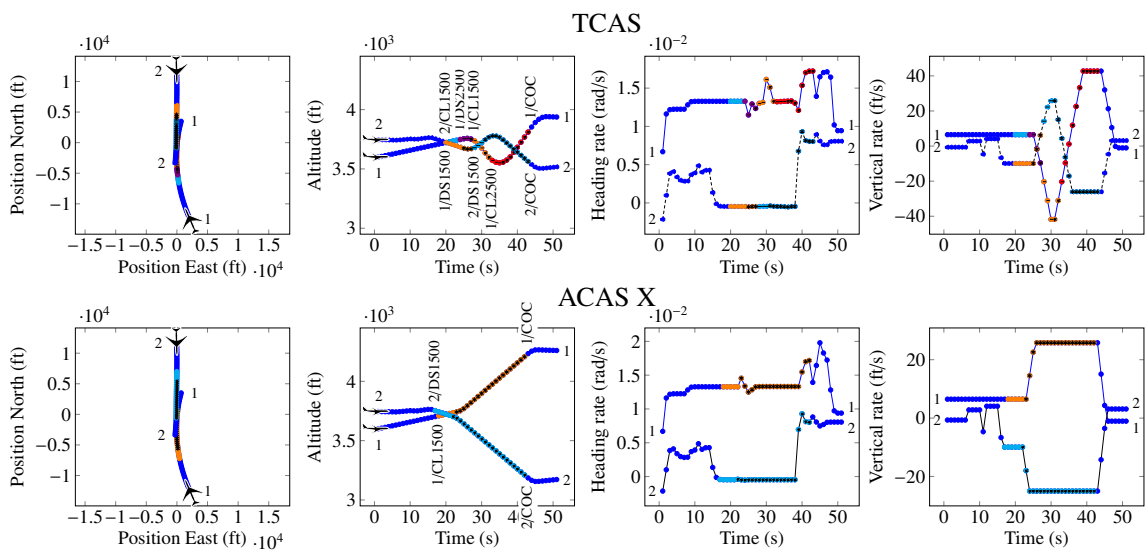}
}
\caption{An NMAC encounter where the aircraft cross in altitude three times.  \Iac{nmac} occurs at 40 seconds in the \ac{tcas} simulation but not in the \ac{acasx} simulation.}
\label{fig:dast2_0590_k1}
\end{figure}

\FloatBarrier
\paragraph{\textbf{Triple Altitude Crossing.}} An example is shown in \Cref{fig:dast2_0590_k1} where \iac{nmac} occurs at 40 seconds in the \ac{tcas} simulation but no \ac{nmac} occurs in the \ac{acasx} simulation.  The encounter occurs with probability density $1.5 \cdot 10^{-9}$, which is more likely than the first two examples presented in this section according to our model.  
The aircraft approach almost straight, head-on, and co-altitude.  \ac{tcas} issues initial \acp{ra} just as the aircraft cross in altitude. Responding to the \acp{ra} actually results in bringing the aircraft closer together vertically.  \ac{tcas} subsequently reverses the \acp{ra}, but the aircraft cross in altitude for a second time due to the pilot response delay.  The aircraft, following the reversal \acp{ra}, cross in altitude for a third time, where \iac{nmac} occurs.  
In contrast, \ac{acasx} is able to resolve the conflict by issuing crossing \acp{ra} slightly earlier than \ac{tcas} and maintaining the \acp{ra} for the duration of the encounter. 

In the \ac{tcas} simulation, interactions between the \ac{ra} and the pilot response delay lead to oscillations in the aircraft altitudes.  In particular, the response delay causes \acp{ra} and their responses to be separated by altitude crossings, resulting in the responses reducing vertical separation rather than increasing it.  The oscillations result in multiple altitude crossings and ultimately \iac{nmac}.   
Another interesting feature of the encounter is the change in vertical rate by aircraft 2 at 17 seconds, which immediately precedes the initial \acp{ra} and the oscillations.  Due to the proximity of the maneuver, the effects of state estimation may be playing a significant role in the initial \ac{ra}. 
Due to hardware limitations, the collision avoidance system does not directly measure the vertical rate of the intruder and must estimate it by tracking altitude over time.  Furthermore, the altitude of the intruder is not known exactly, but quantized to 25 feet increments.  To deal with this uncertainty, tracking and filtering techniques are used, which introduce a small amount of tracking error and lag.  \ac{tcas} uses an alpha-beta filter while \ac{acasx} uses a more sophisticated modified Kalman filter \parencite{Asmar2013}.  The modified Kalman filter has been shown to give better tracking error and lag performances.  The improved state tracking may be a key contributor to the earlier initial \ac{ra} and better selection of the vertical direction in the \ac{acasx} encounter.

\begin{figure}[!htbp]
\centering\resizebox{1.0\columnwidth}{!}
{
\input{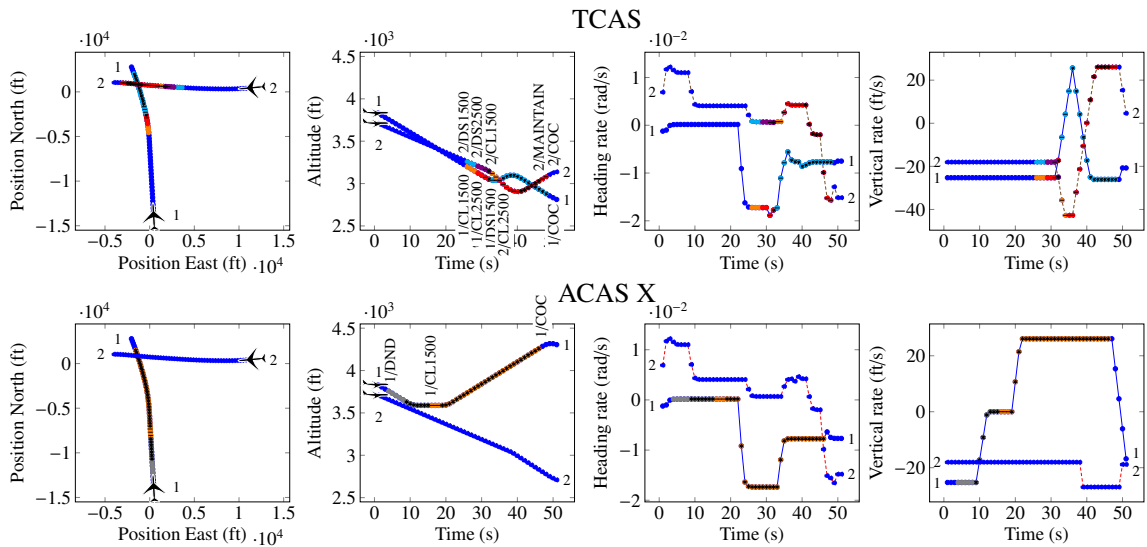}
}
\caption{An NMAC encounter where the aircraft are in a vertical chase configuration.  TCAS issues crossing RAs after the aircraft have crossed in altitude and are beginning to diverge vertically.  \Iac{nmac} occurs at 43 seconds in the \ac{tcas} simulation but not in the \ac{acasx} simulation.}
\label{fig:dast2_0726_k1}
\end{figure}

\paragraph{\textbf{Vertical Chase.}} An example is shown in \Cref{fig:dast2_0726_k1} where \iac{nmac} occurs at 43 seconds in the \ac{tcas} simulation but no \ac{nmac} occurs in the \ac{acasx} simulation.  The encounter occurs with probability density $8.4 \cdot 10^{-9}$, which is more likely than the first two examples presented in this section according to our model.   In the encounter, the aircraft approach perpendicularly in the horizontal direction and are engaged in a vertical chase.  The aircraft are both descending but at different vertical rates.  The aircraft cross in altitude and begin to diverge vertically when \ac{tcas} issues a crossing \ac{ra}.  Responding to the initial \ac{ra}, the aircraft cross in altitude for a second time.  \ac{tcas} then reverses the advisory, causing the aircraft to cross in altitude a third time, where \iac{nmac} occurs.
In contrast, \ac{acasx} resolves the conflict by issuing a preventative DND advisory early followed by a CL1500 to maintain vertical separation of the aircraft.

The \ac{tcas} encounter shows similar oscillations as in the previous example in \Cref{fig:dast2_0590_k1}.  However, there are a couple of key differences.  First, the crossing initial advisories are issued after the aircraft have already crossed in altitude and are beginning to diverge vertically.  Second, this encounter has a change in heading rate immediately preceding the initial \acp{ra} rather than a change in vertical rate.  As a result, it is likely that both vertical state tracking and the estimation of time to closest horizontal approach are playing a role in the initial \acp{ra}.

\FloatBarrier
\subsection{Performance Comparison with Direct Monte Carlo Simulation}
\label{sec:performance}
We compare the performance of \ac{mcts} against direct Monte Carlo sampling given a fixed computational budget.  The algorithms are given a fixed amount of wall clock time and the best path found at the end of that time is returned.  We compare the wall clock time of the algorithms rather than number of samples to account for the additional computations performed in the \ac{mcts} algorithm.  We use the same configuration as the single-threat encounter experiments as shown in \Cref{tab:singlethreatconfig} except that we limit the search based on computation time instead of a fixed number of iterations.  The experiments were performed on a laptop with an Intel i7 4700HQ quad-core processor and 32 GB of memory.

\begin{figure}[!htbp]
	\centering
	\resizebox{0.85\columnwidth}{!}
    {
    \begin{subfigure}[b]{0.5\columnwidth}
        \centering
		\resizebox{1.0\columnwidth}{!}
		{
        	\begin{tikzpicture}[transform shape,font=\large]
\begin{axis}[view = {0}{90}, legend pos = south east, ylabel = Return, xlabel = Computation Time (s)]\addplot+ [,error bars/.cd, x dir=both, x explicit, y dir=both, y explicit]
coordinates {
(250.0, -2055.2046300387065) +=(0.0,167.8487495856387) -=(0.0,167.8487495856387)
(500.0, -1954.9513249528372) +=(0.0,164.64812049742636) -=(0.0,164.64812049742636)
(1000.0, -1899.2140517693108) +=(0.0,164.93102837986095) -=(0.0,164.93102837986095)
(1500.0, -1862.0138397159753) +=(0.0,163.92479468883928) -=(0.0,163.92479468883928)
(2000.0, -1816.2704112777544) +=(0.0,162.37987204991686) -=(0.0,162.37987204991686)
(2500.0, -1741.1196331154313) +=(0.0,157.4026332201176) -=(0.0,157.4026332201176)
};
\addlegendentry{MC}
\addplot+ [,error bars/.cd, x dir=both, x explicit, y dir=both, y explicit]
coordinates {
(250.0, -1818.1324202407156) +=(0.0,185.7363379418811) -=(0.0,185.7363379418811)
(500.0, -1714.363620218265) +=(0.0,186.7558782139564) -=(0.0,186.7558782139564)
(1000.0, -1599.8925953259413) +=(0.0,180.17910477764198) -=(0.0,180.17910477764198)
(1500.0, -1634.461084720134) +=(0.0,189.70022030905588) -=(0.0,189.70022030905588)
(2000.0, -1520.3317725446227) +=(0.0,181.60790722964313) -=(0.0,181.60790722964313)
(2500.0, -1494.654736709784) +=(0.0,180.38658707449824) -=(0.0,180.38658707449824)
};
\addlegendentry{MCTS}
\end{axis}

\end{tikzpicture}
		}
		\centering
		\caption{}
		\label{fig:rvi}
    \end{subfigure}%
    \begin{subfigure}[b]{0.5\columnwidth}
        \centering
		\resizebox{0.915\columnwidth}{!}
		{
			\begin{tikzpicture}[transform shape,font=\large]
\begin{axis}[view = {0}{90}, legend pos = north west, ylabel = NMAC Percentage (\%), xlabel = Computation Time (s)]\addplot+ coordinates {
(250.0, 0.0)
(500.0, 0.0)
(1000.0, 0.0)
(1500.0, 0.0)
(2000.0, 0.0)
(2500.0, 0.0)
};
\addlegendentry{MC}
\addplot+ coordinates {
(250.0, 4.0)
(500.0, 4.0)
(1000.0, 8.0)
(1500.0, 9.0)
(2000.0, 15.0)
(2500.0, 17.0)
};
\addlegendentry{MCTS}
\end{axis}

\end{tikzpicture}
		}
		\centering
		\caption{}
		\label{fig:nmacvi}
    \end{subfigure}
    }
	\caption{Performance of MCTS and Monte Carlo with computation time.  MCTS is able to find an increasing number of \acp{nmac} while Monte Carlo is unable to find any due to the vast search space and rare failure events.}
	\label{fig:nmacperformance}
\end{figure}
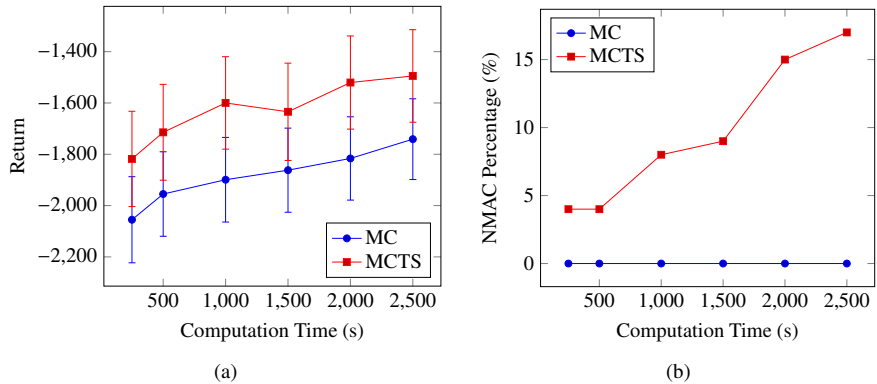

\Cref{fig:nmacperformance} shows the performance of the two algorithms as computation time varies.  \Cref{fig:rvi} compares the return of the best encounters found by the algorithms.  Each data point shows the mean and standard error of the mean of 100 pairwise encounters.  \Cref{fig:nmacvi} shows the \ac{nmac} find rate out of the 100 encounters searched.  In both cases, \ac{mcts} clearly outperforms the baseline Monte Carlo search.  The effectiveness of \ac{mcts} in finding \acp{nmac} is particularly important and we see that \ac{mcts} greatly outperforms the baseline in this regard.  As the computational budget increases, \ac{mcts} is able to find increasingly many \acp{nmac}, whereas at the computational budgets considered, Monte Carlo is unable to find any \acp{nmac}.

\FloatBarrier
\section{Conclusion}
This article presented \acf{ast}, a reinforcement learning-based stress testing approach for finding the most likely path to a failure event.  We described \ac{ast} formulations for the case where the state of the simulator is fully observable and also for the case where the state is hidden.  For the latter case, we presented the \ac{mctssa} algorithm that uses the pseudorandom seed of the simulator to overcome partial observability.  We also presented \acf{dast}, an extension of \ac{ast} for stress testing relative to a baseline.  We applied \ac{ast} and \ac{dast} to stress test a prototype of the next-generation \ac{acasx} in an aircraft encounter simulator and found a number of categories of near mid-air collisions, which we reported to the \ac{acasx} team.  Our differential studies of \ac{acasx} with \ac{tcas} give us additional confidence that \ac{acasx} will offer a significant added safety benefit compared to \ac{tcas}.  Our results contributed to the certification case of \ac{acasx}, which led to the acceptance of \ac{acasx} by the RTCA.
 Our implementation of \ac{ast} is available as an open source Julia package at\\ \url{https://github.com/sisl/AdaptiveStressTesting.jl}.

\acks{We thank Neal Suchy at the Federal Aviation Administration; Wes Olson, Robert Klaus, and Cindy McLain at MIT Lincoln Laboratory; Rachel Szczesiul and Jeff Brush at Johns Hopkins Applied Physics Laboratory; and others on the ACAS X team. We thank Guillaume Brat at NASA Ames Research Center and Corina Pasareanu at Carnegie Mellon University Silicon Valley for their valuable feedback.  We thank Anthony Corso, Robert Moss, Mark Koren, and Rory Lipkis for their useful feedback on the article.  This work was supported by the Safe and Autonomous Systems Operations (SASO) Project and System-Wide Safety (SWS) Project under NASA Aeronautics Research Mission Directorate (ARMD) Airspace Operations and Safety Program (AOSP); and also supported by the Federal Aviation Administration (FAA) Traffic-Alert \& Collision Avoidance System (TCAS) Program Office (PO) AJM-233, Volpe National Transportation Systems Center Contract No. DTRT5715D30011.}


\vskip 0.2in
\printbibliography

\end{document}